\documentclass[letterpaper]{article}
\let\TreePINNPDFInfo\pdfinfo
\usepackage[preprint]{aaai2027}
\usepackage[hyphens]{url}
\usepackage{graphicx}
\usepackage{natbib}
\usepackage{caption}
\usepackage{amsmath,amssymb}
\usepackage{booktabs,array}
\usepackage{algorithm}
\usepackage{algorithmic}
\newcommand{\TreePINNTypeset}{%
  \ifmmode\text{TreePIN\textsuperscript{2}}%
  \else TreePIN\textsuperscript{2}\fi}
\DeclareRobustCommand{\TreePINN}{%
  \ifdefined\texorpdfstring
    \texorpdfstring{\TreePINNTypeset}{TreePIN2}%
  \else\TreePINNTypeset\fi}
\TreePINNPDFInfo{%
  /Title (TreePIN2: Tree-Prior Initialized Neural Networks)
  /Author (Wang Yanrong; Sun Qiming; Haoyang Chen)}
\title{\TreePINN{}: Tree-Prior Initialized Neural Networks}
\author{%
\begin{tabular}[t]{@{}c@{}}
Wang Yanrong\\[.3em]
{\normalfont\normalsize Department of Computer Science}\\
{\normalfont\normalsize National University of Singapore}\\
{\normalfont\normalsize Singapore, Singapore}\\
{\normalfont\normalsize yanrong.wang@u.nus.edu}
\end{tabular}
\hspace{1em}%
\begin{tabular}[t]{@{}c@{}}
Sun Qiming\\[.3em]
{\normalfont\normalsize Department of Computer Science}\\
{\normalfont\normalsize National University of Singapore}\\
{\normalfont\normalsize Singapore, Singapore}\\
{\normalfont\normalsize sqiming@comp.nus.edu.sg}
\end{tabular}
\hspace{1em}%
\begin{tabular}[t]{@{}c@{}}
Haoyang Chen\corresponding\\[.3em]
{\normalfont\normalsize Department of Computer Science}\\
{\normalfont\normalsize National University of Singapore}\\
{\normalfont\normalsize Singapore, Singapore}\\
{\normalfont\normalsize chen\_haoyang@u.nus.edu}
\end{tabular}
}
\affiliations{}
\newcommand{\relu}{\operatorname{ReLU}}


\begin{document}
\maketitle
\raggedbottom
\begingroup
\clubpenalty=10000
\widowpenalty=10000
\brokenpenalty=10000

\begin{abstract}
Fitted decision trees provide predictive knowledge for neural network initialization. We present Tree-Prior Initialized Neural Networks (\TreePINN{}), a method that translates split rules, decision paths and leaf predictions into initial network weights and biases for classification and regression. Sigmoid units encode split responses, hidden units combine them along decision paths, and leaf predictions initialize the output layer. We calibrate split softness using local training margins, sample support and impurity reduction, relating the initial responses to the data at each node. All weights and biases remain trainable, allowing gradient-based optimization to refine the transferred decisions and their combinations. Across six classification and six regression tasks with twenty paired seeds, \TreePINN{} improves mean initial and early validation performance over initialization baselines on most tasks. After fifty epochs, it achieves the best or tied-best mean validation performance among the six compared methods on ten tasks. Paired ablations show how split calibration and leaf-informed output initialization contribute to the starting predictor and its subsequent learning.
\end{abstract}

\section{Introduction}
\label{sec:introduction}

Neural network initialization determines the function from which optimization begins. Common random initializations set parameter scales to support learning, while a fitted decision tree already contains predictive knowledge of the task. Its splits identify useful feature thresholds, its paths combine decisions, and its leaves provide predictions for the resulting regions. Encoding this information in neural parameters gives learning a starting point informed by the relationship between inputs and targets.

Transferring this knowledge requires a representation of the context in which each tree decision is made. A threshold deep in a tree applies to the samples selected by its preceding decisions, and its contribution to prediction depends on the leaves reached afterward. The neural construction must therefore coordinate responses across layers so that local split tests combine into the intended paths and outputs. This connection between local decisions and complete predictions makes the initial function an explicit object of design. By expressing it through trainable weights and biases, the tree's predictive organization becomes a starting representation that neural optimization can develop as a whole.

We propose Tree-Prior Initialized Neural Networks (\TreePINN{}), a method for constructing this starting function from a fitted tree. The construction expresses each split as a pair of sigmoid responses, combines successive decisions through hidden units, and initializes the output layer from leaf predictions. Classification and regression share the same hidden construction, with task-specific output weights. All weights and biases remain trainable, allowing optimization to refine the split tests, their combinations and their predictive associations.

A central design choice is the softness of the initial split responses. Sharp sigmoids approximate hard tree decisions away from the thresholds, where their derivatives become small. Softer responses change over a broader input range, and their departures from hard branch decisions accumulate as conditions are combined along a path. The same numerical slope also produces different responses at nodes whose samples lie at different distances from their thresholds. \TreePINN{} sets a local slope profile using these distances, sample support and impurity reduction. It then selects a common scale by comparing an auxiliary predictor with source-tree predictions on training inputs. The resulting slopes are encoded in the initial weights and biases before neural optimization.

We evaluate the method across twelve tasks with twenty paired seeds, measuring predictive performance at initialization, over the first ten epochs and after fifty epochs. These measurements characterize both the predictive information transferred from the tree and its subsequent refinement through neural optimization. Comparisons with random parameters in the same architecture examine the contribution of the tree-derived initialization; comparisons with the source tree show how the resulting network develops beyond the fitted model. Paired scale and readout ablations identify the roles of calibrated split responses and leaf-informed outputs across these stages. \TreePINN{} achieves the best or tied-best mean validation performance on ten tasks at the fixed endpoint and exceeds its source tree's mean validation performance on eleven tasks after training.

\newpage
\section{Related Work}

\paragraph{Controlling initial signal scales.}
Xavier and He initialization control weight variance to support signal propagation in deep networks~\cite{RP17,RP18}. Data-dependent methods use observed activations or gradients to set initial scales: LSUV calibrates layer-output variance~\cite{RP19}, while data-dependent initialization can balance approximate unit learning rates~\cite{DataDependentInit2016}. Adaptive activation methods learn activation scales during optimization~\cite{AdaptiveActivation2019}. \TreePINN{} sets scales from the supervised decisions of a fitted tree. Training-data margins define the scale around each threshold, while sample support and impurity reduction determine a target sigmoid response. These choices are incorporated into the affine parameters before neural optimization.

\paragraph{Transferring tree structure.}
Tree-derived initializations use fitted models to identify useful structure for neural learning. DJINN uses tree dependencies to construct connectivity and initialize a network~\cite{RP02}, while STBI transfers tree-based feature interactions into sparse neural initializations~\cite{RP32}. Neural Random Forests also connect tree predictors with neural architectures and weights~\cite{NeuralRandomForests2018}. \TreePINN{} initializes a trainable computation from the split thresholds, ordered paths and leaf outputs of a single source tree per run, with split responses calibrated to local training data. Our evaluation compares these initialization methods under a common parameter budget and examines the effects of split calibration and output initialization through paired ablations.

\paragraph{Trainable tree representations.}
Tree-to-network translation also supports editable explanations, where neural representations make tree decisions accessible to inspection and modification~\cite{EditableXAI}. \TreePINN{} addresses predictive initialization by specifying the split responses, path computations and output weights that form the network's starting function. We examine both the predictions supplied by this construction and their development through subsequent optimization.

\section{Technical Approach}
\label{sec:method}

\TreePINN{} encodes a fitted tree's split rules, decision paths and leaf predictions in the initial weights and biases of a neural network. Sigmoid units express the split conditions, hidden units combine conditions along each path, and the output layer converts path activities into predictions. For example, a path requiring both $x_1\leq\tau_1$ and $x_2>\tau_2$ becomes two soft conditions followed by a unit that combines them; its output connection starts from the corresponding leaf prediction. We first construct these computations, then calibrate the split responses from training data and train the resulting network.

Let $\mathcal D_{\mathrm{tr}}=\{(\mathbf x_i,y_i)\}_{i=1}^{n}$ be the training set and $\mathcal T$ a binary tree fitted to it. The tree has $I\geq1$ internal nodes and leaf set $\mathcal L$. At node $v$, it tests feature $j_v$ against threshold $\tau_v$. Branches 0 and 1 denote the left and right children; equality belongs to the left branch.

\subsection{From Tree Decisions to Network Predictions}
\label{sec:tree-prior}

\paragraph{Split rules become sigmoid responses.}
Each split becomes a pair of differentiable responses that retain its feature and threshold. For slope $\alpha_v>0$, the two branch responses are
\begin{equation}
 q_{v,0}(\mathbf x)=\sigma(\alpha_v(\tau_v-x_{j_v})),\quad
 q_{v,1}(\mathbf x)=\sigma(\alpha_v(x_{j_v}-\tau_v)).
 \label{eq:condition}
\end{equation}
Here $\sigma$ is the logistic sigmoid. The responses sum to one at initialization: far to the left, $q_{v,0}$ is close to one and $q_{v,1}$ is close to zero; to the right, the roles reverse. Both equal $1/2$ at the threshold. Larger $\alpha_v$ makes the transition sharper. Each source node receives its own pair of units, including when several nodes test the same feature.

\paragraph{Hidden units combine decisions along a path.}
A leaf is reached when all conditions on its path are satisfied. We express this conjunction by extending the activity $h$ of a partial path with its next condition $q$:
\begin{equation}
 h'=\relu(h+q-1).
 \label{eq:path-step}
\end{equation}
Here $\relu(t)=[t]_+=\max(0,t)$. For binary inputs this acts as an AND: it returns one only when both inputs are one. For soft inputs, two values of $.9$ give activity $.8$; a third $.9$ condition reduces it to $.7$. Each condition therefore contributes to the activity of the complete path. For a leaf $\ell$ with path $P_\ell$ and branch choice $b_\ell(v)$ at node $v$, repeated application gives
\begin{equation}
 h_\ell^{(0)}(\mathbf x)=
 \left[1-\sum_{v\in P_\ell}(1-q_{v,b_\ell(v)}(\mathbf x))\right]_+.
 \label{eq:leaf-path}
\end{equation}
Unit-weight connections carry each condition to the layer where it is needed and carry completed shallow paths to the final hidden layer. Equation~\eqref{eq:leaf-path} specifies the initialized path computation, whose connections are subsequently trainable. Algorithm~\ref{alg:mapping} specifies the complete construction.

\paragraph{Leaf predictions initialize the output layer.}
\label{par:component-H}
The output layer combines path activities using the predictions stored at the leaves. Classification and regression share the hidden construction and use task-specific output coefficients.

For classification with $K$ classes, a path contributes to the score of its leaf's majority class. If leaf $\ell$ predicts class $c_\ell$, the initial logit for class $k$ is
\begin{equation}
 f_k^{(0)}(\mathbf x)=\sum_{\ell:c_\ell=k}h_\ell^{(0)}(\mathbf x).
 \label{eq:classification-head}
\end{equation}
Each original leaf-to-class association has total weight one and the output bias is zero; replicated leaf units share that coefficient equally. Training uses cross-entropy on the logits; softmax converts them to class probabilities.

For regression, a path contributes its leaf's mean target. We standardize targets as $z=(y-\mu_y)/s_y$, using the training mean $\mu_y$ and population SD $s_y$. With standardized leaf mean $c_\ell$ and $H$ hidden layers, the initial scalar output is
\begin{equation}
 f^{(0)}(\mathbf x)=\sum_\ell c_\ell h_\ell^{(0)}(\mathbf x),
 \qquad b_{H+1}^{(0)}=0.
 \label{eq:regression-head}
\end{equation}
Replicated leaf units share the original coefficient equally. Training minimizes mean squared error in $z$, and evaluation restores predictions to the original target units. These output mappings combine the transferred leaf predictions according to the network's soft path activities.

\paragraph{Joint optimization of the initialized network.}
\label{sec:training}
The construction uses a sigmoid first layer, ReLU hidden layers and a linear output. For input dimension $d$, output dimension $o$, and $H$ hidden layers of widths $w_1,\ldots,w_H$, its forward computation is
\begin{align}
 \mathbf a_1&=\sigma\!\left(g_0(\mathbf W_1\mathbf x+\mathbf b_1)\right),\nonumber\\
 \mathbf a_r&=\relu(\mathbf W_r\mathbf a_{r-1}+\mathbf b_r),\quad r=2,\ldots,H,\label{eq:forward}\\
 \mathbf f(\mathbf x)&=\mathbf W_{H+1}\mathbf a_H+\mathbf b_{H+1}.\nonumber
\end{align}
The fixed gain $g_0=20$ defines the first-layer parameterization: each split's affine parameters include the factor $\alpha_v/g_0$, giving effective slope $\alpha_v$ at initialization. The affine coefficients and the operations used to align network dimensions and parameter counts are specified in the supplement. All weights and biases, including entries initialized to zero, are optimized jointly with Adam, beginning with a newly initialized optimizer state.

At initialization, a split unit responds to the single feature selected by its source node. Its full incoming weight vector is trainable, so the response can develop to incorporate additional features. The hidden weights and biases that express path conjunctions are optimized together with the output coefficients that associate paths with predictions. This joint parameterization makes the learned tests, their combinations and the resulting predictions available for refinement under the same task loss.

\subsection{Calibrating the Initial Split Responses}
\label{par:component-S}
\label{sec:auto-alpha}

Split slopes determine both the response around each threshold and the activity propagated along a path. Sharp sigmoids approximate hard decisions away from the threshold and have small derivatives there. Softer responses change more gradually and accumulate larger deficits along the source-tree path through Eq.~\eqref{eq:leaf-path}. The same numerical slope also produces different responses at nodes whose samples lie at different distances from their thresholds. \TreePINN{} calibrates each slope using the training samples reaching that node, then selects a common multiplier using the source tree's predictions.

\paragraph{Node-specific response calibration.}
For node $v$, we estimate a characteristic distance from its threshold using the median nonzero distances on the two branches. We shrink this estimate toward a tree-wide reference according to sample count, giving margin $\widetilde M_v$. We also set a target sigmoid response $\kappa_v$ from sample support and impurity reduction, assigning larger targets to splits supported by more samples and greater impurity reduction. The initial slope profile is
\begin{equation}
 a_v=C\!\left(\operatorname{logit}(\kappa_v)/\widetilde M_v\right),
 \quad C(t)=\operatorname{clip}(t,.5,150).
 \label{eq:auto-profile}
\end{equation}
Before clipping, this choice gives $\sigma(a_v\widetilde M_v)=\kappa_v$. The slope therefore expresses the target response at a distance characteristic of that node's training data. The margin estimator, response targets and numerical safeguards are detailed in the supplement.

The margin and response target play complementary roles. The margin expresses the distance scale on which a node's sigmoid is evaluated. A smaller margin calls for a steeper slope to reach the same target response, while a larger margin permits a gentler transition. Sample support and impurity reduction determine how strongly the response favors the selected branch at that characteristic distance. Shrinking local margins toward the tree-wide reference ties the distance estimate to the available data: well-supported nodes retain more of their local scale, while nodes with fewer observations draw more strongly on the reference. The common-scale search then evaluates how these locally defined responses combine across the tree.

\paragraph{Calibration against source-tree predictions.}
The local profile sets the relative scales of the split responses. We choose a common multiplier by evaluating their combined predictions against the source tree on training inputs. An auxiliary predictor combines the sigmoid conditions into product path masses. For classification, these masses weight source-leaf log probabilities, and the criterion compares the predicted classes with the source classes. For regression, they weight source-leaf means, and the criterion measures prediction error relative to the spread of source predictions. Classification additionally tests whether a softer profile retains the required agreement. Algorithm~\ref{alg:auto-init} specifies the finite searches and fallback rules. All slope selection uses training data.

The scale search uses normalized product-path routing, with source-leaf log probabilities supplying class-distribution information for the classification proxy. The selected slopes then initialize the network's sigmoid split units. Its initial predictions follow the rectified path computation and majority-class or leaf-mean readout in Eqs.~\eqref{eq:leaf-path}--\eqref{eq:regression-head}, after which all corresponding weights and biases are optimized jointly.

\section{Evaluation}
\label{sec:evaluation}

We evaluate how tree-derived initialization affects predictions before optimization, early learning, and performance after a common training budget. The benchmark compares six methods across twelve tasks. Comparisons with source trees and random initialization of the same architecture examine the value of the transferred parameters, while paired ablations examine split calibration and leaf-informed output initialization.

\begin{table*}[t]\centering\small\setlength{\tabcolsep}{3pt}\renewcommand{\arraystretch}{1.04}
\caption{Validation performance at epoch 50.}\label{tab:init-main}\label{tab:init-class-main}\label{tab:init-reg-main}
\noindent\textbf{Classification: Macro-F1 (\%) $\uparrow$}\par\smallskip
\begin{tabular*}{\textwidth}{@{\extracolsep{\fill}}lrrrrrr@{}}\toprule
Initialization & Bank & Banknote & Breast Cancer & Digits & Spambase & Wine\\\midrule
Random & $71.85\pm1.51$ & $99.94\pm.18$ & $95.81\pm1.71$ & $96.23\pm.93$ & $92.87\pm.98$ & $70.27\pm20.19$\\
He & $71.61\pm1.86$ & $\mathbf{99.96\pm.16}$ & $95.52\pm1.65$ & $95.67\pm1.20$ & $92.73\pm1.09$ & $73.28\pm16.53$\\
Xavier & $\mathit{71.86\pm1.93}$ & $\mathit{99.96\pm.11}$ & $96.00\pm1.98$ & $\mathit{96.44\pm1.03}$ & $93.09\pm.80$ & $\mathit{81.44\pm13.37}$\\
STBI & $71.44\pm2.33$ & $99.32\pm.67$ & $94.46\pm1.19$ & $89.64\pm1.91$ & $\mathbf{93.63\pm.64}$ & $75.38\pm14.83$\\
DJINN & $71.18\pm1.67$ & $99.67\pm.48$ & $\mathit{96.45\pm1.67}$ & $88.98\pm3.36$ & $93.42\pm.93$ & $42.99\pm16.44$\\
\addlinespace[2pt]
\TreePINN{} & $\mathbf{72.62\pm1.62}$ & $\mathbf{99.96\pm.16}$ & $\mathbf{96.66\pm1.35}$ & $\mathbf{97.20\pm.74}$ & $\mathit{93.44\pm.97}$ & $\mathbf{96.78\pm3.10}$\\
\bottomrule\end{tabular*}
\par\medskip
\noindent\textbf{Regression: RMSE $\downarrow$}\par\smallskip
\begin{tabular*}{\textwidth}{@{\extracolsep{\fill}}lrrrrrr@{}}\toprule
Initialization & \shortstack{California\\$10^5$ USD} & \shortstack{Diabetes\\score} & \shortstack{Airfoil\\dB} & \shortstack{Concrete\\MPa} & \shortstack{Protein\\RMSD} & \shortstack{Gas CO\\mg/m$^3$}\\\midrule
Random & $.56\pm.02$ & $\mathit{54.66\pm3.23}$ & $2.85\pm.39$ & $6.23\pm.67$ & $4.20\pm.07$ & $1.23\pm.11$\\
He & $.56\pm.02$ & $56.05\pm3.97$ & $2.72\pm.38$ & $6.15\pm.63$ & $\mathit{4.09\pm.07}$ & $1.18\pm.13$\\
Xavier & $\mathit{.56\pm.02}$ & $55.08\pm3.87$ & $\mathit{2.69\pm.37}$ & $\mathit{5.99\pm.59}$ & $4.12\pm.06$ & $1.24\pm.13$\\
STBI & $.56\pm.02$ & $55.25\pm3.68$ & $2.85\pm.32$ & $6.31\pm.46$ & $4.21\pm.08$ & $\mathit{1.17\pm.10}$\\
DJINN & $.63\pm.03$ & $\mathbf{53.83\pm3.42}$ & $4.27\pm.22$ & $8.37\pm.87$ & $4.66\pm.07$ & $1.23\pm.09$\\
\addlinespace[2pt]
\TreePINN{} & $\mathbf{.53\pm.02}$ & $65.43\pm4.86$ & $\mathbf{2.12\pm.24}$ & $\mathbf{5.54\pm.55}$ & $\mathbf{4.03\pm.07}$ & $\mathbf{1.15\pm.11}$\\
\bottomrule\end{tabular*}
\par\smallskip\begin{minipage}{\textwidth}\small Mean $\pm$ sample SD over twenty paired seeds. Boldface and italics indicate the best and second-best distinct unrounded means; statistical comparisons are reported in Table~\ref{tab:corrected-main}. On Banknote, \TreePINN{} and He tie, and Xavier has a nearly identical mean.\end{minipage}
\end{table*}

\begin{figure*}[t]
\centering
\includegraphics[width=\textwidth]{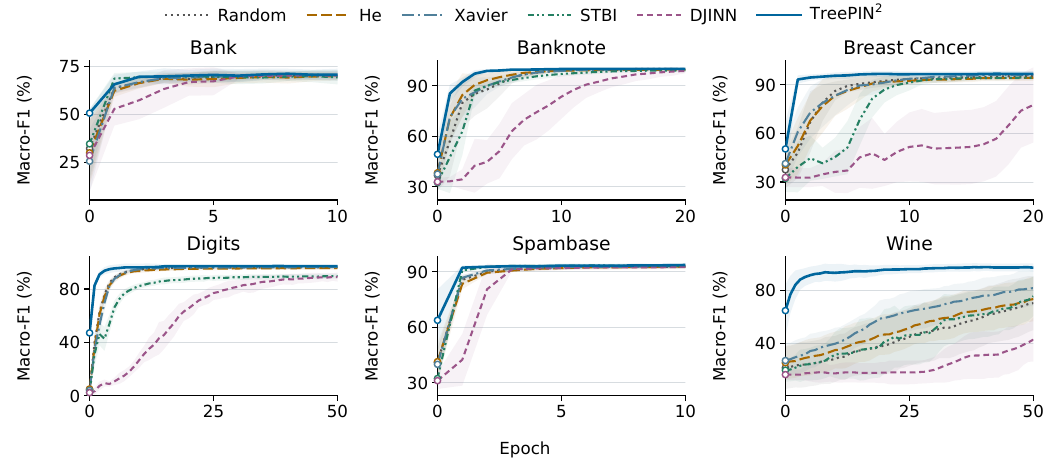}
\caption{Classification validation Macro-F1 over twenty paired seeds per method; curves show the mean and shading indicates one sample SD. Open circles mark epoch 0 before optimization. Bank and Spambase show epochs 0--10, Breast Cancer and Banknote 0--20, and Digits and Wine 0--50, with task-specific linear vertical ranges. Full epochs 0--50 appear in the supplement.}
\label{fig:init-class-curves}
\par
\includegraphics[width=\textwidth]{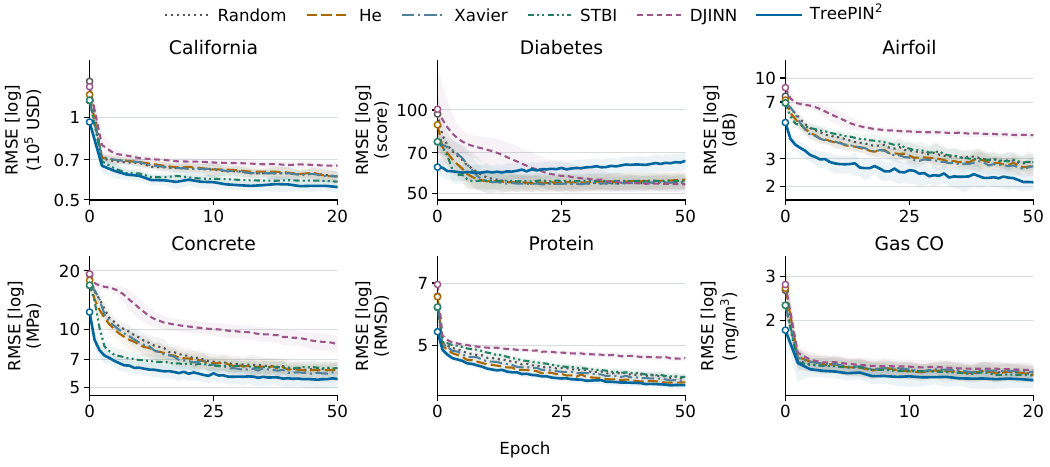}
\caption{Regression validation RMSE over twenty paired seeds per method; curves show the mean and shading indicates one sample SD. Open circles mark epoch 0 before optimization. California and Gas show epochs 0--20; Airfoil, Diabetes, Protein and Concrete show 0--50. Logarithmic vertical axes show initial and later errors in each target's units. Full epochs 0--50 on linear axes appear in the supplement.}
\label{fig:init-reg-curves}
\end{figure*}

\begin{table*}[t]\centering\small\setlength{\tabcolsep}{3pt}\renewcommand{\arraystretch}{1.04}

\begingroup\centering\small\setlength{\tabcolsep}{3pt}\renewcommand{\arraystretch}{1.04}
\caption{Scale and readout ablation at epoch 50.}\label{tab:init-components-main}
\begin{tabular*}{\textwidth}{@{\extracolsep{\fill}}lrrrrr@{}}\toprule
Initialization & \shortstack{Digits\\F1 (\%) $\uparrow$} & \shortstack{Spambase\\F1 (\%) $\uparrow$} & \shortstack{Concrete\\RMSE (MPa) $\downarrow$} & \shortstack{Protein\\RMSE (RMSD) $\downarrow$} & \shortstack{Gas CO\\RMSE (mg/m$^3$) $\downarrow$}\\\midrule
\shortstack[l]{Without adaptive scales\\and leaf readout} & $94.266\pm1.339$ & $\mathit{93.439\pm.820}$ & $\mathit{5.518\pm.418}$ & $4.104\pm.056$ & $\mathbf{1.143\pm.073}$\\
Without adaptive scales & $94.466\pm1.201$ & $93.414\pm.743$ & $5.596\pm.503$ & $4.109\pm.071$ & $1.173\pm.101$\\
Without leaf readout & $\mathit{96.909\pm.676}$ & $93.309\pm1.213$ & $\mathbf{5.414\pm.562}$ & $\mathbf{4.021\pm.085}$ & $1.148\pm.093$\\
\addlinespace[2pt]
\TreePINN{} & $\mathbf{97.196\pm.742}$ & $\mathbf{93.441\pm.965}$ & $5.541\pm.546$ & $\mathit{4.029\pm.066}$ & $\mathit{1.146\pm.107}$\\
\bottomrule\end{tabular*}
\par\smallskip\begin{minipage}{\textwidth}\small Mean $\pm$ sample SD over twenty paired seeds. All variants share the hidden architecture and mini-batch sequence. Removing adaptive scales fixes every split slope at 20; removing the leaf-informed readout replaces it with a nonzero random matrix of equal Frobenius norm. Boldface and italics indicate the best and second-best distinct unrounded means.\end{minipage}
\endgroup
\par\medskip
\caption{Budgeted optimization cost as a percentage of the 50-epoch update budget. Bold and italics indicate the lowest and second-lowest mean costs.}\label{tab:updates-target}
\noindent\textbf{Classification}\par\smallskip
\begin{tabular*}{\textwidth}{@{\extracolsep{\fill}}lrrrrrr@{}}\toprule
Initialization & Bank & Banknote & Breast Cancer & Digits & Spambase & Wine\\\midrule
Random & $\mathit{21.3\pm27.9}$ & $\mathit{29.4\pm17.4}$ & $47.4\pm31.3$ & $51.4\pm33.7$ & $22.5\pm27.0$ & $83.1\pm30.5$\\
He & $29.7\pm30.7$ & $41.1\pm14.4$ & $51.2\pm25.6$ & $54.5\pm22.1$ & $16.2\pm6.3$ & $92.6\pm8.7$\\
Xavier & $22.6\pm25.1$ & $34.4\pm23.0$ & $\mathit{38.2\pm30.0}$ & $\mathit{49.6\pm34.9}$ & $18.1\pm20.2$ & $\mathit{68.7\pm36.9}$\\
STBI & $33.6\pm35.0$ & $79.7\pm28.7$ & $69.2\pm38.9$ & $100.0\pm0.0$ & $\mathit{12.2\pm22.2}$ & $73.1\pm30.9$\\
DJINN & $45.5\pm40.4$ & $61.6\pm23.8$ & $57.2\pm20.2$ & $100.0\pm0.0$ & $26.0\pm27.0$ & $96.2\pm9.4$\\
\addlinespace[2pt]
\TreePINN{} & $\mathbf{12.2\pm11.6}$ & $\mathbf{13.3\pm4.1}$ & $\mathbf{12.4\pm21.2}$ & $\mathbf{16.4\pm20.6}$ & $\mathbf{9.0\pm11.6}$ & $\mathbf{4.8\pm7.0}$\\
\bottomrule\end{tabular*}
\par\medskip
\noindent\textbf{Regression}\par\smallskip
\begin{tabular*}{\textwidth}{@{\extracolsep{\fill}}lrrrrrr@{}}\toprule
Initialization & California & Diabetes & Airfoil & Concrete & Protein & Gas CO\\\midrule
Random & $84.4\pm23.3$ & $30.4\pm26.6$ & $88.2\pm20.8$ & $78.2\pm23.8$ & $97.9\pm5.6$ & $63.9\pm36.5$\\
He & $76.9\pm16.0$ & $\mathbf{24.9\pm20.8}$ & $80.6\pm21.9$ & $79.2\pm19.8$ & $\mathit{82.4\pm10.8}$ & $\mathit{53.3\pm32.8}$\\
Xavier & $\mathit{68.9\pm17.3}$ & $\mathit{27.1\pm23.2}$ & $\mathit{73.1\pm25.7}$ & $\mathit{65.9\pm23.0}$ & $94.1\pm8.5$ & $71.6\pm37.1$\\
STBI & $74.6\pm32.7$ & $29.6\pm33.6$ & $88.2\pm17.5$ & $75.7\pm34.4$ & $98.6\pm5.8$ & $55.4\pm38.2$\\
DJINN & $99.6\pm1.8$ & $54.1\pm23.6$ & $100.0\pm0.0$ & $100.0\pm0.0$ & $100.0\pm0.0$ & $80.0\pm36.1$\\
\addlinespace[2pt]
\TreePINN{} & $\mathbf{27.9\pm15.6}$ & $58.4\pm47.3$ & $\mathbf{29.6\pm15.4}$ & $\mathbf{34.9\pm30.1}$ & $\mathbf{68.8\pm16.8}$ & $\mathbf{27.3\pm22.1}$\\
\bottomrule\end{tabular*}
\par\smallskip\begin{minipage}{\textwidth}\small Mean $\pm$ sample SD over twenty paired seeds; lower values indicate fewer updates ($\downarrow$). Cost stops at first observed target attainment or budget exhaustion. Nonattaining runs incur 100\% of the budget. Across 240 runs per method, the numbers reaching the target are 164 for Random, 240 for He, 195 for Xavier, 122 for STBI, 95 for DJINN and 224 for \TreePINN{}. Table~\ref{tab:updates-target-attainment} reports attainment counts and censored medians by task.\end{minipage}
\end{table*}

\subsection{Experimental Setup}

\paragraph{Tasks and data.}
The benchmark includes six classification tasks---Bank Marketing, Banknote, Breast Cancer, Digits, Spambase and Wine---and six regression tasks---California Housing, Diabetes, Airfoil, Concrete, Protein Structure and Gas Turbine CO. The main comparisons and core ablations use twenty paired seeds. Additional experiments evaluate Diamonds, Airbnb Berlin and Superconductor with three seeds. Table~\ref{tab:init-datasets} lists dataset sources, target units and preprocessing.

All preprocessing, source trees and calibration statistics are fitted on training data. Reported results use validation partitions; reserved test partitions were not evaluated. Gas validation measures within-period performance on a random split of the 2011--2013 development data, with 2014--2015 reserved for test. Airbnb uses disjoint listing groups. The supplement gives the full split and preprocessing specifications.

\paragraph{Compared initializations.}
We compare Random, He, Xavier, STBI~\cite{RP32}, DJINN~\cite{RP02} and \TreePINN{} under common dense dimensions and training budgets. Random, He and Xavier use ReLU hidden layers and linear outputs. Random draws Gaussian weights and biases; He uses Kaiming-normal hidden weights, linear-output gain 1 and zero biases; Xavier uses gain-1 uniform weights and zero biases throughout. STBI uses Tanh activations, DJINN uses ReLU, and \TreePINN{} uses the sigmoid/ReLU/linear computation in Eq.~\eqref{eq:forward}. STBI and DJINN use the cited tree-to-network constructions, adapted to single trees and shared dimensions as detailed in the supplement.

\paragraph{Matching network dimensions.}
Within each dataset and seed, methods share the input data, fitted source tree, mini-batch order and dense network dimensions. For classification, an identity ReLU layer after DJINN's first hidden layer aligns its depth with the \TreePINN{} mapping, and each common hidden width is the maximum of the two aligned widths. Replication copies a unit's incoming weights and bias and divides its outgoing coefficients among the copies. The copies therefore make the same combined contribution to the next layer at initialization, preserving the mapped function in the wider network. Identity ReLU connections carry nonnegative hidden activities through an additional layer when depth alignment requires it. Regression uses the compact \TreePINN{} dimensions, with mapped baseline blocks embedded through zero padding and identity connections. The supplement specifies these alignment operations. The parameter budget in Eq.~\eqref{eq:actual-parameters} counts all stored weights and biases, including initial zeros. The matched networks retain the patterns of initial activity and replica symmetry induced by each method's construction.

\paragraph{Training and metrics.}
All networks train for 50 epochs with Adam, learning rate $.001$, betas $(.9,.999)$, epsilon $10^{-8}$ and zero weight decay. Batch sizes are 128 for classification and 64 for regression. \TreePINN{} and its ablations use the fixed first-layer gain $g_0=20$. Source trees have maximum depth 5 and minimum leaf size 5, except for Bank, whose settings appear in the supplement.

The primary metrics are validation Macro-F1 in percent for classification and validation RMSE in the target's units for regression. We evaluate performance before optimization, average performance over epochs 1--10 within each run, and performance at the common endpoint of epoch 50. Classification cross-entropy measures predictive loss. Results are summarized by means and sample SDs across the repeated data splits; paired comparisons use the corresponding seed-level differences.

\subsection{Main Results}

\TreePINN{} achieves the best or tied-best mean validation performance on ten of twelve tasks after fifty epochs (Table~\ref{tab:init-main}). It has the highest mean Macro-F1 on Bank, Breast Cancer, Digits and Wine. On Banknote, its mean Macro-F1 ties that of He and is close to Xavier's. STBI leads on Spambase, with $93.63\%$ compared with \TreePINN{}'s $93.44\%$.

\begin{samepage}
The advantage extends to regression, where \TreePINN{} has the lowest mean RMSE on California, Airfoil, Concrete, Protein and Gas. Diabetes is the exception: its mean RMSE is $65.43$, compared with $56.05$ for He and $53.83$ for DJINN. Table~\ref{tab:init-paired} reports the paired differences relative to He across all twelve tasks.
\end{samepage}

\subsection{Learning Dynamics}

The learning curves show the value of the informed starting function: \TreePINN{} improves mean initial and early validation performance over He on all twelve tasks (Figures~\ref{fig:init-class-curves} and~\ref{fig:init-reg-curves}). Bank and Spambase improve most rapidly in the first few epochs, while Digits continues improving throughout training. In regression, the early advantages on Airfoil and Concrete persist through epoch 50, with smaller final improvements on California and Protein. Diabetes follows a different trajectory: its initial advantage reverses during training. On Breast Cancer, higher final Macro-F1 accompanies higher cross-entropy, distinguishing the gains in class decisions from predictive loss (Figure~\ref{fig:init-class-config-loss-full}). The supplement provides the complete trajectories from epoch 0 to 50.

The comparison with random initialization of the same architecture confirms the predictive value of the transferred parameters. \TreePINN{} improves mean initial performance on all twelve tasks and mean early performance on eleven, with Bank as the exception (Table~\ref{tab:same-architecture-random}). At epoch 50, Holm-adjusted comparisons identify advantages on Airfoil, Concrete and Protein; the classification differences are not statistically significant.

Neural optimization develops the transferred representation beyond its source tree: after fifty epochs, \TreePINN{} achieves better mean validation performance on eleven of twelve tasks. These gains emerge during training, with the initialized network starting below the tree on eleven tasks. Diabetes has final RMSE $65.43$, compared with $64.18$ for the tree. Tables~\ref{tab:source-tree-classification} and~\ref{tab:source-tree-regression} report the tree and network results together with initial transfer agreement or normalized prediction error.

\paragraph{Updates to a shared performance target.}
Across twelve tasks and twenty paired seeds, we count optimizer updates to the first observed attainment of the corresponding He baseline's epoch-50 validation performance. Attainment is evaluated at initialization and at each epoch end. Table~\ref{tab:updates-target} reports the update cost through first attainment or budget exhaustion; runs that do not reach the target incur the full budget. Kaplan--Meier median update counts are lower for \TreePINN{} than for He on eleven tasks. On Diabetes, nine of twenty \TreePINN{} runs reach the target, so the median is not reached within the budget (Table~\ref{tab:updates-target-attainment}). The update count measures neural optimization, with tree fitting and initialization contributing additional computation; it does not measure elapsed runtime.

\paragraph{Paired statistical analysis.}
Across 180 corrected-resampled comparisons with Holm adjustment, \TreePINN{} performs significantly better in 23/60 comparisons at initialization, 39/60 over epochs 1--10, and 12/60 at epoch 50. At epoch 50, significant advantages occur on two tasks against Random, four against STBI and six against DJINN; no comparison against He or Xavier is significant. Table~\ref{tab:init-main} presents the corresponding performance means, and Tables~\ref{tab:corrected-main} and~\ref{tab:corrected-efficiency} report the adjusted performance and optimization-cost comparisons.

\subsection{Ablation Study}
\begin{samepage}

We examine split calibration and leaf-informed output initialization in a paired two-by-two study on Digits, Spambase, Concrete, Protein and Gas. Each condition uses twenty paired seeds. The four variants share the sigmoid/ReLU/linear network, dense dimensions, optimizer and batch sequence, varying the split scales and output initialization.

\end{samepage}

\paragraph{Adaptive scales.}
We compare the calibration procedure with a fixed effective slope $\alpha_{\mathrm{fixed}}=20$ at every split. The remaining hidden construction and corresponding readout are shared. This comparison evaluates the combined use of local margins, sample support, impurity reduction and the common scale search.

Adaptive calibration improves early learning and final performance on Digits. Mean epoch-50 F1 increases by $2.73$ percentage points with the leaf-informed readout and $2.64$ points with the random readout, with all twenty paired differences favoring calibration in both settings (Table~\ref{tab:init-components-main}). Mean cross-entropy over the first ten epochs also decreases under both readouts (Figure~\ref{fig:init-calibration-curves}). These gains show that split calibration improves learning across the two output initializations.

Calibration also lowers mean final RMSE on Concrete and Protein under both readouts. On Gas, it lowers mean final RMSE with the leaf-informed readout and raises it with the random readout. Spambase shows little change in mean final F1 and higher early cross-entropy with adaptive scales. A separate ten-seed study extends the scale comparison to all six classification tasks (Table~\ref{tab:init-ablation}).

We also assess slope assignment by permuting slopes within matching signed-replica groups while preserving the grouped slope distribution. Across ten seeds, this permutation produces small, variable performance differences on Digits and Spambase (Table~\ref{tab:init-assignment}).

\paragraph{Leaf-informed readout.}
The readout ablation replaces the leaf-informed output matrix with a gain-1 Xavier-uniform matrix rescaled to the same Frobenius norm. The output bias remains zero, and each scale condition retains its corresponding hidden initialization. The adaptive and fixed-scale variants use the same random matrix. Full readout specifications are provided in the supplement.

Leaf-informed output initialization provides an immediate predictive advantage on all five ablation tasks with adaptive scales. On Digits, it also improves mean epoch-50 F1 under both scale settings. In regression, random readouts reach lower mean final RMSE in five of the six combinations of task and scale setting. These results establish the immediate predictive contribution of the leaf-informed readout and show that its effect on final performance depends on the task and split-scale setting (Tables~\ref{tab:init-components} and~\ref{tab:init-component-pairs}).

\subsection{Discussion}

\TreePINN{} gives neural optimization a starting function informed by the decisions learned in a fitted tree. Split thresholds, path connections and leaf outputs encode task-specific structure before any gradient update. The improvements in initial predictions and early learning under the same architecture demonstrate the value of transferring this structure through the initial parameters.

The Digits ablation shows how the benefit of split calibration develops during learning. With the leaf-informed readout, adaptive scales reduce mean initial F1 from $64.62\%$ under fixed scales to $47.15\%$, yet raise mean early F1 from $88.29\%$ to $93.83\%$ and epoch-50 F1 from $94.47\%$ to $97.20\%$ (Table~\ref{tab:init-components}). The gain in final F1 also persists with a random readout. Calibration therefore improves the learning trajectory on Digits even though the starting predictor is less accurate.

Leaf-informed output weights contribute immediate predictive information, improving mean initial performance on all five ablation tasks with adaptive scales. Subsequent training refines the split tests, path combinations and output associations together. This adaptation extends the value of the tree prior beyond its initial predictions: after fifty epochs, \TreePINN{} exceeds its source tree's mean validation performance on eleven of twelve tasks. The construction thus combines a predictive prior from the fitted tree with a neural representation that can be refined through learning.

\section{Conclusion}
\TreePINN{} initializes neural networks from fitted tree decisions and calibrates their split responses using local training margins, sample support and impurity reduction. All weights and biases remain trainable. Across six classification and six regression tasks with twenty paired seeds, it improves mean initial and early validation performance over initialization baselines on most tasks and achieves the best or tied-best mean validation performance on ten tasks after fifty epochs. The ablations establish the initial predictive value of leaf-informed outputs and the contribution of split calibration to subsequent learning. The method makes a fitted tree's predictive knowledge the starting point for joint refinement of its decisions and their combinations.

\par
\endgroup


\clearpage

\appendix
\section{Data and Experimental Setup}
\label{sec:init-full-panel}

\paragraph{Data sources and paired splits.}
Table~\ref{tab:init-datasets} lists the fifteen datasets, their sources and target units. UCI identifiers refer to the official repository~\cite{UCIRepository2023}. The six main classification and six main regression tasks each use twenty paired seeds. The five classification tasks other than Bank use stratified 60/20/20 partitions with seeds 0--19 and the same seed at both split stages. Bank uses grouped stratified five-fold partitions with seeds 100--119: fold 0 forms the test set, fold 1 the validation set, and folds 2--4 the training set. Identical feature rows are grouped before splitting. Main regression tasks use seeds 100--119, reserving 20\% of observations for testing with seed $s$ and 20\% of the remaining observations for validation with $s+10000$. Gas reserves 2014--2015 for testing and randomly divides the 2011--2013 development data into 80/20 training and validation partitions; its validation results therefore measure within-period performance. All reported evaluations use validation partitions; reserved test partitions are not evaluated.

\begin{table*}[t]\centering
\begingroup\small\setlength{\tabcolsep}{4pt}\renewcommand{\arraystretch}{1.1}
\caption{Datasets, sources and target units.}\label{tab:init-datasets}
\begin{tabular*}{\textwidth}{@{\extracolsep{\fill}}l>{\raggedright\arraybackslash}p{2.2in}rrl>{\raggedright\arraybackslash}p{1.15in}@{}}\toprule
Task & Source & $N$ & $d$ & Split & Target / units\\\midrule
\multicolumn{6}{@{}l}{\textbf{Classification}}\\[2pt]
Bank & UCI 222 (full dataset) & 45,211 & 16 & G & subscription, 2 classes\\[3pt]
Banknote & UCI 267 & 1,372 & 4 & C & authenticity, 2 classes\\[3pt]
Breast Cancer & Wisconsin Diagnostic Breast Cancer\newline(scikit-learn) & 569 & 30 & C & diagnosis, 2 classes\\[3pt]
Digits & scikit-learn Digits & 1,797 & 64 & C & digit, 10 classes\\[3pt]
Spambase & UCI 94 & 4,601 & 57 & C & spam, 2 classes\\[3pt]
Wine & scikit-learn Wine & 178 & 13 & C & cultivar, 3 classes\\[3pt]
\addlinespace[4pt]
\multicolumn{6}{@{}l}{\textbf{Main regression}}\\[2pt]
California & scikit-learn California Housing & 20,640 & 8 & R & median value, $10^5$ USD\\[3pt]
Diabetes & scikit-learn Diabetes\newline(unscaled inputs) & 442 & 10 & R & progression score\\[3pt]
Airfoil & UCI 291 & 1,503 & 5 & R & sound pressure, dB\\[3pt]
Concrete & UCI 165 (official numeric data) & 1,030 & 8 & R & strength, MPa\\[3pt]
Protein & UCI 265 (CASP) & 45,730 & 9 & R & first-column RMSD\\[3pt]
Gas CO & UCI 551, five annual datasets\newline(2011--2015) & 36,733 & 9 & Y & CO, mg/m$^3$\\[3pt]
\addlinespace[4pt]
\multicolumn{6}{@{}l}{\textbf{Supplementary regression}}\\[2pt]
Diamonds & Data accompanying STBI~\cite{RP32} & 53,940 & 9 & R & $\log(1+\mathrm{price}/\mathrm{USD})$\\[3pt]
Airbnb Berlin & STBI listings and calendar data~\cite{RP32} & 119,268 & 13 & L & price, USD\\[3pt]
Superconductor & UCI 464 & 21,263 & 81 & R & critical temperature, K\\[3pt]
\addlinespace[4pt]
\bottomrule\end{tabular*}\par\smallskip\begin{minipage}{\textwidth}\small\textit{Definitions.} $N$ is the number of observations before splitting, and $d$ is the input dimension. Split proportions are given as training/validation/test: R, random rows (approximately 64/16/20); C, stratified 60/20/20; G, grouped stratified 60/20/20; Y, a test set reserved by year and randomly split development data; L, listing-ID grouped 64/16/20. Reserved test partitions are not evaluated.\end{minipage}\endgroup
\end{table*}

\paragraph{Preprocessing and source trees.}
All paired methods share input data, transformations, split indices and source trees. For the five classification tasks other than Bank, min--max scaling is followed by centering and population-SD scaling, with both transformations fitted on training data and zero SDs replaced by 1. Trees are fitted in min--max coordinates, and their thresholds are transformed to the standardized network coordinates. Class labels are integer encoded. Bank uses numeric standardization fitted on training data and categorical codes. For the main regression tasks, inputs are min--max scaled using training statistics, without clipping validation values, and targets are standardized using the training mean and population SD. Protein uses column 1 (RMSD) as the target and columns 2--10 as predictors. Gas excludes NOX and year from predictors; seed 100 has 17,752 training, 4,439 validation and 14,542 reserved test rows. Source trees use maximum depth 5 and minimum leaf size 5, except Bank, which uses depth 4 and minimum support $\max(2,\lceil .005n\rceil)$.

Preprocessing uses scikit-learn and NumPy, and neural training uses PyTorch on Linux CPUs. Additional experiments evaluate Diamonds, Airbnb Berlin and Superconductor with three seeds (Table~\ref{tab:init-reg-exploratory}). Their sources and target units are included in Table~\ref{tab:init-datasets}: Diamonds retains the supplied log-transformed target, Airbnb uses disjoint listing-ID groups, and Superconductor uses critical-temperature units. The two datasets obtained from STBI use the data accompanying that method~\cite{RP32}.

\paragraph{Baseline initialization and network dimensions.}
He and Xavier use ReLU hidden layers. He uses Kaiming-normal hidden weights with ReLU gain $\sqrt2$, linear-output gain 1, and zero biases. Xavier uses gain-1 uniform weights and zero biases throughout. Random draws every weight independently from a zero-mean Gaussian with variance $3/(\mathrm{fan\_in}+\mathrm{fan\_out})$ and each affine bias independently with variance $3/(o+w_H)$, where $o$ is output dimension and $w_H$ the last hidden width. Its random generator uses seed $314159+s$ for classification and $20261019+s$ for regression and is reset after network construction. Parameters are initialized directly in the network's input coordinates, without absorbing a preprocessing transformation into the first affine layer.

STBI and DJINN are evaluated as single-tree initializations with common network dimensions and a fixed 50-epoch training budget. Their original ensemble settings, training budgets and hyperparameter searches are replaced by this shared evaluation protocol. For classification, alignment begins with the compact \TreePINN{} dimensions $[d,2I,\ldots,2I,o]$, inserts one identity ReLU layer after DJINN's first hidden layer, and takes the componentwise maximum of the aligned hidden widths. Widening preserves the initial functions of \TreePINN{} and DJINN in exact arithmetic, and STBI is embedded in the same dimensions. Regression uses the compact \TreePINN{} dimensions, with native STBI and DJINN blocks embedded through zero padding and, for DJINN, identity connections before the readout. The construction requires each mapping to be compatible with the target dimensions. All affine entries, including those initialized to zero, count toward the stored parameter budget.

Each of the 240 dataset--seed combinations uses fixed input data, split indices, fitted preprocessing transformations, source-tree parameters and network dimensions across methods. The same inputs and fitted transformations are reused throughout each paired comparison without refitting preprocessing.

\section{Mapping and Calibration}
\label{sec:parameter-calibration-details}
\label{sec:supplement}

The forward computation, path activities and task readouts are defined in Eqs.~\eqref{eq:forward}, \eqref{eq:leaf-path}, \eqref{eq:classification-head} and~\eqref{eq:regression-head}. The construction applies to trees with at least one internal node. Internal nodes are ordered by a depth-first traversal and then stably sorted by feature, retaining each node's identity even when distinct nodes specify numerically identical tests. In Algorithm~\ref{alg:mapping}, $\mathbf A_r=\mathbf W_{r+1}$ and $\mathbf c_r=\mathbf b_{r+1}$ denote the affine weights and biases after the first split layer. Identity connections preserve conditions awaiting combination and completed leaf paths; regression uses the same hidden construction with a scalar readout from standardized leaf means.

\paragraph{First-layer parameter coordinates.}
If $s(v)$ indexes the negative condition row, its nonzero initial entries and those of its complement are
\begin{align}
 W^{(0)}_{1,s(v),j_v}&=-\alpha_v/g_0,&b^{(0)}_{1,s(v)}&=\alpha_v\tau_v/g_0,\nonumber\\
 W^{(0)}_{1,s(v)+1,j_v}&=\alpha_v/g_0,&b^{(0)}_{1,s(v)+1}&=-\alpha_v\tau_v/g_0.
 \label{eq:condition-parameters}
\end{align}
The fixed gain $g_0=20$ defines the first-layer parameterization throughout training. Each calibrated slope is encoded through the factor $\alpha_v/g_0$; substituting the parameters into Eq.~\eqref{eq:forward} cancels the gain at initialization and yields the effective slope $\alpha_v$ in Eq.~\eqref{eq:condition}. Calibration therefore sets the initial decision softness within a common parameterization. Adam updates the stored weights and biases directly. Changing the gain while compensating the initial parameters preserves the initial split responses but can alter subsequent training at the same learning rate.

\paragraph{Depth, widths and replicas.}
Let $D$ be the maximum number of split decisions on a tree path, with $H\geq D$, $w_0=d$ and $w_{H+1}=o$. The compact mapping has $D$ hidden layers of width $2I$. Identity ReLU layers before the readout align depth. Width alignment retains the original units and assigns additional units by cycling deterministically through eligible source units. Eligible units have a nonzero incoming affine and a nonzero outgoing column; after the sigmoid layer, analytically inactive ReLUs with all incoming coefficients and bias nonpositive are excluded. Incoming rows and biases are copied, and outgoing coefficients are divided by copy multiplicity. Leaf-readout coefficients are likewise divided among final-port replicas. These operations preserve the initialized function in exact arithmetic. The mapping requires target dimensions and eligible source units that support the specified alignment. Calibration assigns one effective slope per original source node to all its replicas. The stored parameter count is
\begin{equation}
 P=\sum_{r=1}^{H+1}(w_{r-1}+1)w_r.
 \label{eq:actual-parameters}
\end{equation}
All weights and biases, including initial zeros, are available to the optimizer. With zero weight decay and the implemented derivative at zero, inactive ReLU units can retain zero gradients, while exact replicas can maintain identical gradients and Adam states in exact arithmetic. Matching tensor dimensions controls the stored parameter budget, while effective capacity at initialization also depends on activation patterns and replica symmetry.

\paragraph{Initialized path activity.}
To derive the initialized path activity, write $[z]_+=\max(0,z)$. For $d\geq0$,
$[[z]_+-d]_+=[z-d]_+$. Substituting $d=1-q$ proves Eq.~\eqref{eq:leaf-path} by induction: the first condition has activity $q_1$; the second gives $[q_1+q_2-1]_+$; and each later condition adds its own nonnegative deficit $1-q$. A shallow leaf's identity connection preserves its nonnegative activity without increasing path length. The identity is exact in real arithmetic for the initialized unit-weight connections and specified biases; subsequent optimization updates these parameters. At a threshold, finite sigmoid conditions assign one-half to both branches.

\begin{algorithm}[H]
\small
\caption{Tree-derived split templates and path connections}
\label{alg:mapping}
\begin{algorithmic}[1]
\REQUIRE Nonconstant tree $\mathcal T$, target shape $(d,w_1,\ldots,w_H,o)$, target-to-source replica maps $\pi_r$
\STATE Build compact shape $[d,2I,\ldots,2I,o]$ ($D$ hidden layers); initialize all coefficients and biases to zero
\STATE Assign $s(v)$ by stable feature sorting of the DFS internal-node list
\STATE Write signed split templates: $\mathbf t_{s(v)}=-\mathbf e_{j_v}$, $c_{s(v)}=\tau_v$ and complementary positive row
\STATE Call $\textsc{Connect}(\mathrm{root},0)$
\STATE Insert $H-D$ identity ReLU layers immediately before the readout
\STATE For each target hidden unit $i$, copy source unit $\pi_r(i)$; let $m_{r,k}=|\{i:\pi_r(i)=k\}|$, with identity input/output maps and input multiplicities 1
\STATE Align every affine by $W'_{r,i,j}=W_{r,\pi_r(i),\pi_{r-1}(j)}/m_{r-1,\pi_{r-1}(j)}$, $b'_{r,i}=b_{r,\pi_r(i)}$; apply the same maps to split templates and leaf ports
\STATE Return aligned templates, path connections, leaf-port maps and readout multiplicities for initialization and optimization in Algorithm~\ref{alg:auto-init}
\end{algorithmic}
\smallskip
\textbf{Procedure} $\textsc{Connect}(v,t)$ on the compact construction, with $\mathbf A_r=\mathbf W_{r+1}$ and $\mathbf c_r=\mathbf b_{r+1}$:
\begin{algorithmic}[1]
\FOR{each child $c$ of $v$, using parent port $p=s(v)+b$}
\IF{$c$ is a leaf}
 \STATE Set $A_{r,p,p}=1$ in intermediate affine layers $r=t+1,\ldots,D-1$
 \STATE Record leaf port $p$ and its task-specific leaf prediction for the readout
\ELSE
 \FOR{$u\in\{s(c),s(c)+1\}$}
  \STATE Set $A_{t+1,u,p}=1$ and $c_{t+1,u}=-1$
  \STATE Set $A_{r,u,u}=1$ for $r=1,\ldots,t+1$
 \ENDFOR
 \STATE Call $\textsc{Connect}(c,t+1)$
\ENDIF
\ENDFOR
\end{algorithmic}
\end{algorithm}

\paragraph{Local margins and target responses.}
Local margins use the training samples reaching each source node. Let $m_{v,L}$ and $m_{v,R}$ be the median nonzero absolute distances to $\tau_v$ on the two sides, and set $M_v=\max(10^{-6},(m_{v,L}+m_{v,R})/2)$. Empty sides use the pooled nonzero median, with fallback 1. For a node reached by $n_v$ training samples, the adjusted margin is
\begin{equation}
 \widetilde M_v=\exp\!\left(c+\frac{n_v}{n_v+a}(\log M_v-c)\right),
 \label{eq:auto-margin}
\end{equation}
where $c=\operatorname{median}_v\log M_v$ and $a=\max(\operatorname{median}_v n_v,1)$.

For normalized weighted support $s_v$ and nonnegative impurity decrease $G_v$, let
\begin{equation}
 \rho_v=\sqrt{\frac{s_v}{\max_u s_u+10^{-12}}
                    \frac{G_v}{\max_u G_u+10^{-12}}}.
 \label{eq:auto-reliability}
\end{equation}
Let $R_v$ be its normalized zero-based rank, using average ranks for ties and $R_v=.5$ when $I=1$. The adjusted reliability is $\bar\rho_v=\operatorname{clip}(\rho_v+.15(R_v-.5),0,1)$. With a task-dependent response floor $\kappa_{\min}$,

\begin{equation}
 \kappa_v=\kappa_{\min}+(.99-\kappa_{\min})\bar\rho_v.
 \label{eq:auto-confidence}
\end{equation}
Together with Eq.~\eqref{eq:auto-profile}, these quantities define the local scale profile; the task-dependent floors and common-scale criteria follow.

\paragraph{Classification calibration.}
\label{sec:classification}
The calibration proxy uses product path masses $\widetilde w_\ell=\prod_{v\in P_\ell}q_{v,b_\ell(v)}$.
Classification calibration uses $u_K=1/\sqrt{\max(1,K-1)}$, $\kappa_{\min}=.75+.10u_K$ and agreement floor $F_K=.55+.10u_K$. The proxy incorporates the source leaf's empirical class probabilities: for probability $p_{\ell,k}$ at leaf $\ell$, its logits are
\begin{equation}
 \widetilde f_k(\mathbf x;\boldsymbol\alpha)=
 \sum_\ell\widetilde w_\ell(\mathbf x;\boldsymbol\alpha)
 \log\max(p_{\ell,k},10^{-6}).
 \label{eq:auto-proxy}
\end{equation}
The common scale search compares the proxy argmax with the hard source-leaf class on training inputs and accepts agreement at least $F_K$. Let $A$ be the selected base profile's measured agreement. To reduce sharpness while retaining part of this agreement, classification calibration uses the target
\begin{equation}
 T=F_K+.5(A-F_K).
 \label{eq:auto-target}
\end{equation}
The softening search first uses at most 2,048 ordered training rows, then verifies the selected clipped profile against $T$ on all training inputs. Failed verification returns the measured base profile.

\paragraph{Regression calibration.}
\label{sec:regression}
Regression uses $\kappa_{\min}=.85$ and the shared margin/reliability profile, with proxy $\widetilde f(\mathbf x)=\sum_\ell\widetilde w_\ell(\mathbf x)c_\ell$. For hard source prediction $t(\mathbf x)$ in standardized target units, the passing criterion is
\begin{equation}
 \frac{\sqrt{n^{-1}\sum_i(\widetilde f(\mathbf x_i)-t(\mathbf x_i))^2}}
 {\max(\operatorname{SD}_{\mathrm{tr}}(t),10^{-6})}\leq .45,
 \label{eq:auto-reg-fidelity}
\end{equation}
where the denominator is the population SD of source predictions. Regression returns the selected base profile with no additional softening stage.

\paragraph{Proxy and network predictions.}
Calibration evaluates a product-path proxy, while the initialized network uses the rectified path activities and task-specific readout defined in the main text. Product masses sum to one across leaves, whereas the activities in Eq.~\eqref{eq:leaf-path} need not. Although $\widetilde w_\ell\geq h_\ell^{(0)}$ for each leaf, this activity bound does not order the predictions, which also depend on the task-specific output coefficients. The proxy criteria characterize agreement or error of the calibration predictor and do not bound the actual network's path error or derivative magnitudes. We therefore evaluate initial transfer and subsequent neural learning directly.

\begin{algorithm}[H]
\small
\caption{\TreePINN{} initialization and learning}
\label{alg:auto-init}\label{alg:auto-alpha}
\begin{algorithmic}[1]
\REQUIRE Training data $\mathcal D_{\mathrm{tr}}$; nonconstant tree $\mathcal T$;
\item[]\hspace{\algorithmicindent} task, gain $g_0$, shape and replica maps; regression
\item[]\hspace{\algorithmicindent} target moments estimated on training data
\ENSURE Initial parameters and a trained predictor
\STATE Construct signed templates $(\mathbf t_r,c_r)$, paths
\item[]\hspace{\algorithmicindent} and leaf-port maps (Algorithm~\ref{alg:mapping})
\STATE Initialize the readout from source-leaf predictions,
\item[]\hspace{\algorithmicindent} using standardized leaf means for regression;
\item[]\hspace{\algorithmicindent} divide coefficients among port replicas
\STATE Estimate margins $\widetilde M_v$ and reliability $\bar\rho_v$
\IF{classification with $K$ classes}
\STATE $u_K\gets1/\sqrt{\max(1,K-1)}$
\STATE $\kappa_{\min}\gets .75+.10u_K$; $F_K\gets .55+.10u_K$
\STATE Use the log-leaf proxy and agreement criterion
\ELSE
\STATE Standardize targets; $\kappa_{\min}\gets .85$
\STATE Use the leaf-mean proxy and normalized RMSE
\item[]\hspace{\algorithmicindent} criterion
\ENDIF
\STATE $C(t)\gets\operatorname{clip}(t,.5,150)$
\STATE $\kappa_v\gets\kappa_{\min}+(.99-\kappa_{\min})\bar\rho_v$
\STATE $a_v\gets C(\operatorname{logit}(\kappa_v)/\widetilde M_v)$
\STATE Test $s=1,2,4,8$ on training inputs
\STATE Refine a passing bracket 16 times;
\item[]\hspace{\algorithmicindent} retain $s=1$ if no candidate passes
\STATE $\boldsymbol\alpha^{\mathrm{base}}\gets\operatorname{float32}(C(s\mathbf a))$
\IF{classification}
\STATE Select a profile by the finite softening search;
\item[]\hspace{\algorithmicindent} verify Eq.~\eqref{eq:auto-target} on all training inputs
\STATE $\boldsymbol\alpha\gets$ verified profile, or $\boldsymbol\alpha^{\mathrm{base}}$ on failure
\ELSE
\STATE $\boldsymbol\alpha\gets\boldsymbol\alpha^{\mathrm{base}}$
\ENDIF
\FOR{each signed/replica row $r$ of source node $v$}
\STATE $\mathbf W_{1,r}\gets\operatorname{float32}(\alpha_v/g_0)\mathbf t_r$
\STATE $b_{1,r}\gets\operatorname{float32}(\alpha_v/g_0)c_r$
\ENDFOR
\STATE Initialize Adam and optimize all weights and biases:
\item[]\hspace{\algorithmicindent} cross-entropy for classification;
\item[]\hspace{\algorithmicindent} standardized-target MSE for regression
\end{algorithmic}
\end{algorithm}

\paragraph{Finite calibration searches and fallback rules.}
\label{sec:auto-alpha-details}

Calibration and network initialization use the same training coordinates and source tree. Thresholds and transferred slopes are stored in float32, while reached-side distances, product masses and proxy logits are computed in float64. Depth-first ordering, stable feature sorting and width-alignment maps retain each source node's identity, including distinct nodes with identical numerical thresholds.

To compute node reliability, we first obtain the impurity gain from raw impurity $J_v$ and weighted count $N_v$:
\begin{equation}
 G_v=\max\!\left(0,J_v-\frac{N_{v,L}}{N_v}J_{v,L}
                         -\frac{N_{v,R}}{N_v}J_{v,R}\right).
\end{equation}
The normalized support is $s_v=N_v/N_{\mathrm{root}}$. Numerical safeguards use $10^{-12}$ for node denominators and 1 for the root count; these safeguards leave the positive-count fitted trees in this study unchanged. Node reliability is clipped to $[0,1]$ before applying the tied-rank contrast. Leaf probabilities in the proxy are normalized source values, or uniform for an empty leaf, and clipped at $10^{-6}$ inside the log without renormalizing the clipped vector. Hard proxy routes use left $\leq$ and right $>$ with the stored float32 thresholds. Because routing uses these stored thresholds, source-library evaluation can differ for samples near a threshold.

The base-profile search selects a common scale using a fixed sequence of checks. It tests scale 1 first, followed on failure by scales 2, 4, and 8 until a passing value is found. The most recent failing scale and first passing scale define a bracket for 16 iterations; each passing midpoint replaces the upper bound and each failing midpoint replaces the lower bound. If all four fail, the procedure resets to scale 1 and measures the returned base agreement again. Runs using fallback profiles remain included in the evaluation.

The classification softening search begins with a deterministic subset: all training rows when their count is at most 2,048, or integer indices from an equally spaced sequence spanning row 0 through the last row otherwise. Let $A_S$ be the subset base agreement and $T_S=F_K+.5(A_S-F_K)$. The selected scale starts at 1. If $A_S\geq F_K$, the subset search scans the nine scales $.25,.34375,\ldots,1$ in increasing order. If the first scale passes $T_S$, the procedure selects $.25$ directly. Otherwise, it refines the first failing/passing grid bracket for eight iterations and retains its passing upper bound. If no scale passes, or the subset base is below $F_K$, it retains scale 1 for the next check.

The check on all training rows verifies the subset choice by evaluating the selected profile against $T$ from Eq.~\eqref{eq:auto-target}. On failure, it performs eight refinements between the selected scale and 1 and re-evaluates the retained upper bound. If that profile still fails, it returns the unsoftened, already measured base profile. Attainment of the base agreement floor and the final softening target is assessed separately, including cases in which softening follows an unsuccessful base-floor search. These finite checks evaluate each candidate empirically, without assuming monotone multiclass agreement or guaranteed satisfaction of the target.

\clearpage

\begin{figure*}[p]
\centering
\includegraphics[width=\textwidth]{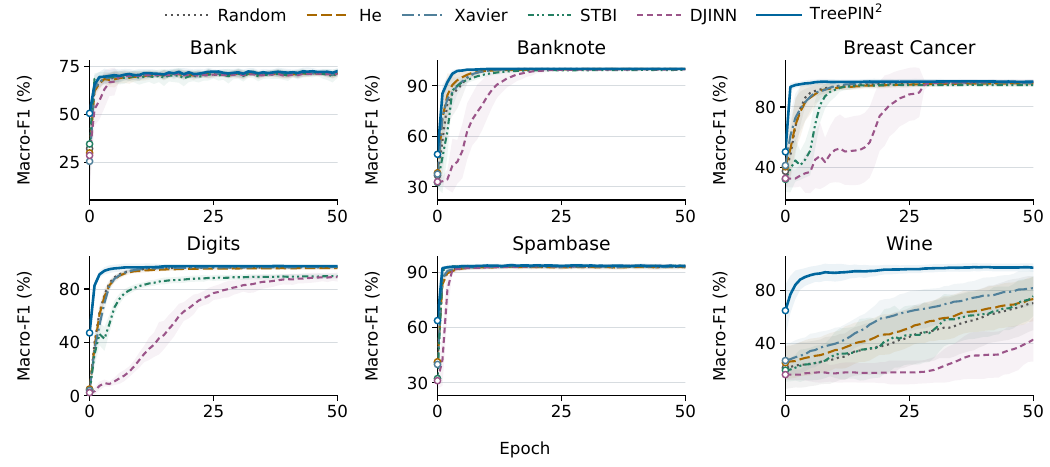}
\caption{Classification learning trajectories for all six methods over epochs 0--50. Curves show mean validation Macro-F1 (\%), with shading for one sample SD over twenty paired seeds. Methods share input data and training budgets and use their specified activation functions. Method styles and the epoch range are consistent across tasks; vertical ranges are task-specific.}
\label{fig:init-class-full}
\end{figure*}

\twocolumn[{\begin{minipage}{\textwidth}
\section{Learning Dynamics and Component Ablations}
\centering
\includegraphics[width=\textwidth]{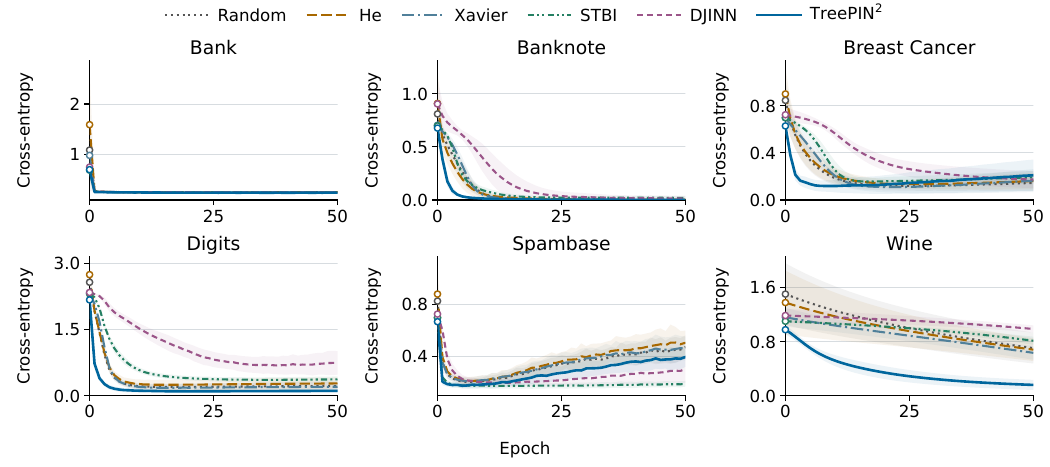}
\captionof{figure}{Classification validation cross-entropy over epochs 0--50 for the six main initialization methods. Curves show the mean and shading indicates one sample SD over twenty paired seeds. Cross-entropy measures predictive loss alongside the Macro-F1 trajectories in Figure~\ref{fig:init-class-full}.}
\label{fig:init-class-config-loss-full}
\end{minipage}\vspace{8pt}}]

\paragraph{Experimental budgets and performance summaries.}
The main comparison comprises 1,440 training trajectories: six methods evaluated on twelve tasks with twenty paired seeds. The five-task, four-variant factorial ablation comprises 400 trajectories, including 100 \TreePINN{} runs shared with the main comparison; these shared runs do not constitute additional independent replications. All primary comparisons use performance at epoch 50. Early performance is averaged over epochs 1--10 within each seed before aggregation across seeds, and sample SD uses denominator $n-1$. Learning curves connect all 51 unsmoothed observations at integer epochs 0--50, with epoch 0 preceding optimization and shading representing variation across seeds. Because repeated splits share source observations, SDs and favorable-pair counts characterize variation across runs. Tables~\ref{tab:corrected-main} and~\ref{tab:corrected-efficiency} report corrected paired inference.

\begin{figure*}[p]
\centering
\includegraphics[width=\textwidth]{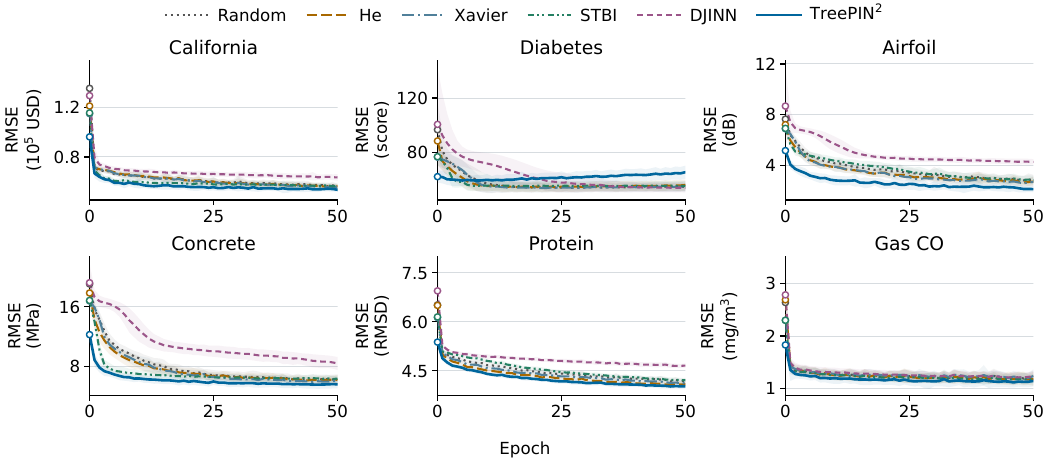}
\caption{Regression learning trajectories for all six methods on the six main tasks over epochs 0--50. Curves show mean validation RMSE in the original target units, with shading for one sample SD over twenty paired seeds. Table~\ref{tab:init-reg-exploratory} reports results on three additional regression tasks.}
\label{fig:init-reg-full}
\end{figure*}

Tables~\ref{tab:init-class-full}, \ref{tab:init-reg-full} and~\ref{tab:init-paired} report initial, early and final performance together with paired differences. Figures~\ref{fig:init-class-full} and~\ref{fig:init-reg-full} show the complete learning trajectories, and Figure~\ref{fig:init-class-config-loss-full} reports classification cross-entropy over the same training budget. On Diabetes, the initial advantage over He reverses by the end of training.

\paragraph{Scale and readout ablations.}
The five-task factorial ablation uses twenty paired seeds per task. The fixed-scale variant sets every effective slope to $\alpha_{\mathrm{fixed}}=20$, preserving the remaining hidden construction and corresponding readout. Random readouts use a CPU PyTorch generator with a deterministic seed for each dataset--seed pair. A SHA-256 digest is computed from a fixed prefix followed by the dataset identifier and split seed, with slash separators. Its first eight bytes are interpreted as a little-endian integer and masked to $2^{63}-1$. The generator draws a gain-1 Xavier-uniform output matrix, which is rescaled by the ratio of the leaf-informed matrix's Frobenius norm to its own norm. The output bias remains zero, and the adaptive and fixed-scale variants use exactly the same random output weights and bias.

The readout ablation changes the output weights' supervised associations and sparsity while holding their Frobenius norm fixed. The root mean square of the outputs and replica symmetry may also change, so the comparison measures the combined effect of this readout replacement. All stored parameters are subsequently optimized.

Within each scale condition, variants with random and leaf-informed readouts share their initial hidden parameters, optimizer settings and mini-batch order. All four variants use the same forward computation and differ in their initial parameters and predictions.

\begin{figure*}[p]
\centering
\includegraphics[width=\textwidth]{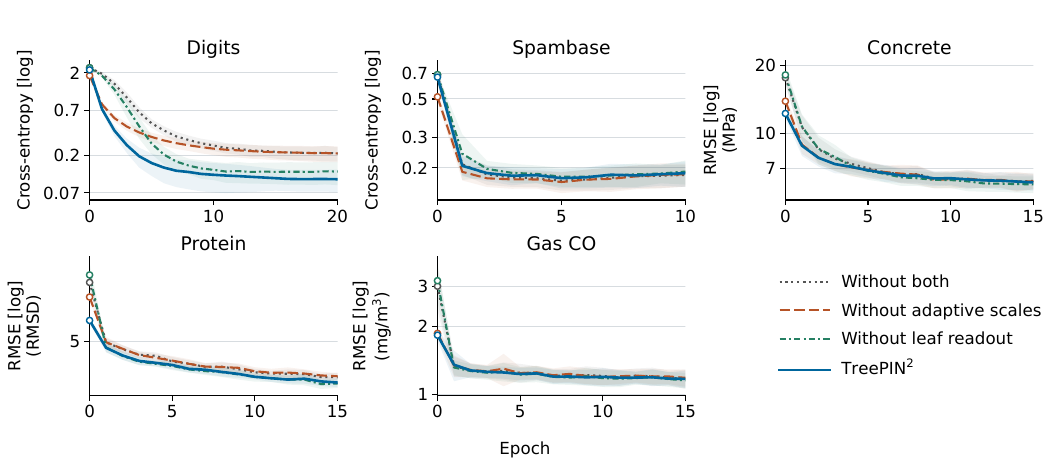}
\caption{Learning dynamics of the four scale and readout combinations in Table~\ref{tab:init-components-main}. Curves show mean validation cross-entropy for classification and RMSE for regression, with shading for one sample SD over twenty paired seeds. Open circles mark epoch 0 before optimization. Digits shows epochs 0--20, Spambase 0--10, and regression tasks 0--15; vertical axes are logarithmic. Curves connect the unsmoothed observations.}
\label{fig:init-calibration-curves}
\end{figure*}

\begin{figure*}[p]
\centering
\includegraphics[width=\textwidth]{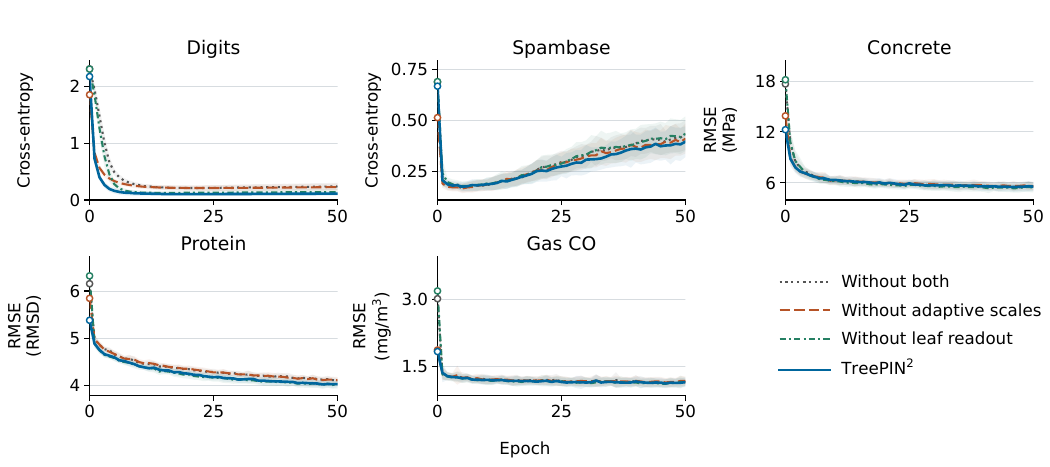}
\caption{Complete scale-by-readout validation trajectories over epochs 0--50. Curves show the mean and shading indicates one sample SD over twenty paired seeds per task and variant. Digits and Spambase use cross-entropy; Concrete, Protein and Gas use RMSE in their stated target units. All vertical axes are linear.}
\label{fig:init-core-full}
\end{figure*}

Tables~\ref{tab:init-components} and~\ref{tab:init-component-pairs} report the four variants and their conditional paired effects. Figure~\ref{fig:init-calibration-curves} shows the corresponding learning windows, and Figure~\ref{fig:init-core-full} reports the complete trajectories through epoch 50. Table~\ref{tab:init-ablation} extends the scale comparison to all six classification tasks using ten seeds.

\paragraph{Node-assignment diagnostic.}
The ten-seed diagnostic in Table~\ref{tab:init-assignment} permutes slopes within matching signed-replica groups. The permutation preserves the grouped slope and signed-weight multisets, while threshold-scaled bias multisets generally change. On Digits and Spambase, the small mean improvements in epoch-50 F1 are smaller than the variability of the paired differences, leaving the contribution of node-specific slope assignment unresolved. The main scale ablation evaluates the slope distribution and node assignment together.

\clearpage

\begin{table*}[p]\centering\small
\setlength{\tabcolsep}{2.5pt}
\renewcommand{\arraystretch}{1.10}

\caption{Classification validation Macro-F1 (\%) for \TreePINN{} and He at initialization, averaged over epochs 1--10, and at epoch 50. Values are mean $\pm$ sample SD over twenty paired seeds; early performance is averaged within each seed before aggregation. Boldface and italics indicate the best and second-best distinct unrounded means within each task and stage; exact ties share rank. Higher Macro-F1 is better. Table~\ref{tab:init-main} compares all six methods at epoch 50.}
\label{tab:init-class-full}
\begin{tabular*}{\textwidth}{@{\extracolsep{\fill}}lrrrrrr@{}}\toprule
 & \multicolumn{2}{c}{Initial} & \multicolumn{2}{c}{Early (1--10)} & \multicolumn{2}{c}{Epoch 50}\\
\cmidrule(lr){2-3}\cmidrule(lr){4-5}\cmidrule(lr){6-7}
Task & \TreePINN{} & He & \TreePINN{} & He & \TreePINN{} & He\\\midrule
Bank & $\mathbf{50.616\pm1.882}$ & $\mathit{29.935\pm16.958}$ & $\mathbf{69.877\pm1.182}$ & $\mathit{68.039\pm1.100}$ & $\mathbf{72.624\pm1.617}$ & $\mathit{71.611\pm1.860}$\\
Banknote & $\mathbf{49.275\pm11.479}$ & $\mathit{38.263\pm11.124}$ & $\mathbf{97.098\pm1.078}$ & $\mathit{92.397\pm4.334}$ & $\mathbf{99.963\pm0.165}$ & $\mathbf{99.963\pm0.165}$\\
Breast Cancer & $\mathbf{50.389\pm16.807}$ & $\mathit{40.523\pm12.381}$ & $\mathbf{95.643\pm1.581}$ & $\mathit{82.141\pm7.916}$ & $\mathbf{96.664\pm1.352}$ & $\mathit{95.517\pm1.646}$\\
Digits & $\mathbf{47.154\pm8.980}$ & $\mathit{5.113\pm2.379}$ & $\mathbf{93.828\pm2.623}$ & $\mathit{80.264\pm2.272}$ & $\mathbf{97.196\pm0.742}$ & $\mathit{95.665\pm1.197}$\\
Spambase & $\mathbf{63.795\pm16.727}$ & $\mathit{41.371\pm10.205}$ & $\mathbf{93.216\pm0.945}$ & $\mathit{91.074\pm0.797}$ & $\mathbf{93.441\pm0.965}$ & $\mathit{92.730\pm1.093}$\\
Wine & $\mathbf{64.700\pm19.388}$ & $\mathit{25.768\pm12.530}$ & $\mathbf{89.347\pm6.842}$ & $\mathit{30.498\pm13.041}$ & $\mathbf{96.784\pm3.098}$ & $\mathit{73.281\pm16.528}$\\
\bottomrule\end{tabular*}
\end{table*}

\begin{table*}[p]\centering\small
\setlength{\tabcolsep}{2.5pt}
\renewcommand{\arraystretch}{1.10}

\caption{Regression validation RMSE in target units for \TreePINN{} and He at initialization, averaged over epochs 1--10, and at epoch 50. Values are mean $\pm$ sample SD over twenty paired seeds; early performance is averaged within each seed before aggregation. Boldface and italics indicate the best and second-best distinct unrounded means within each task and stage; exact ties share rank. Lower RMSE is better. Table~\ref{tab:init-main} compares all six methods at epoch 50.}
\label{tab:init-reg-full}
\begin{tabular*}{\textwidth}{@{\extracolsep{\fill}}lrrrrrr@{}}\toprule
 & \multicolumn{2}{c}{Initial} & \multicolumn{2}{c}{Early (1--10)} & \multicolumn{2}{c}{Epoch 50}\\
\cmidrule(lr){2-3}\cmidrule(lr){4-5}\cmidrule(lr){6-7}
Task & \TreePINN{} & He & \TreePINN{} & He & \TreePINN{} & He\\\midrule
California ($10^5$ USD) & $\mathbf{0.963\pm0.034}$ & $\mathit{1.213\pm0.172}$ & $\mathbf{0.605\pm0.020}$ & $\mathit{0.675\pm0.014}$ & $\mathbf{0.534\pm0.016}$ & $\mathit{0.561\pm0.022}$\\
Diabetes (score) & $\mathbf{62.152\pm4.429}$ & $\mathit{88.405\pm17.740}$ & $\mathbf{59.864\pm3.839}$ & $\mathit{61.763\pm5.391}$ & $\mathit{65.435\pm4.858}$ & $\mathbf{56.049\pm3.972}$\\
Airfoil (dB) & $\mathbf{5.157\pm0.380}$ & $\mathit{7.203\pm0.651}$ & $\mathbf{3.256\pm0.277}$ & $\mathit{4.557\pm0.280}$ & $\mathbf{2.117\pm0.245}$ & $\mathit{2.715\pm0.383}$\\
Concrete (MPa) & $\mathbf{12.261\pm0.508}$ & $\mathit{17.881\pm1.848}$ & $\mathbf{7.029\pm0.448}$ & $\mathit{10.333\pm0.817}$ & $\mathbf{5.541\pm0.546}$ & $\mathit{6.147\pm0.625}$\\
Protein (RMSD) & $\mathbf{5.380\pm0.045}$ & $\mathit{6.502\pm0.374}$ & $\mathbf{4.593\pm0.042}$ & $\mathit{4.696\pm0.045}$ & $\mathbf{4.029\pm0.066}$ & $\mathit{4.091\pm0.073}$\\
Gas CO (mg/m$^3$) & $\mathbf{1.830\pm0.098}$ & $\mathit{2.688\pm0.588}$ & $\mathbf{1.241\pm0.081}$ & $\mathit{1.316\pm0.094}$ & $\mathbf{1.146\pm0.107}$ & $\mathit{1.183\pm0.132}$\\
\bottomrule\end{tabular*}
\end{table*}

\begin{table*}[p]\centering\small
\setlength{\tabcolsep}{2.5pt}
\renewcommand{\arraystretch}{1.10}

\caption{Paired validation differences between \TreePINN{} and He, reported as \TreePINN{} minus He: mean $\pm$ sample SD over twenty paired seeds. Classification differences are Macro-F1 percentage points (pp), with positive values favoring \TreePINN{}; regression differences are RMSE in target units, with negative values favoring \TreePINN{}. Early performance is averaged over epochs 1--10 within each seed. He uses ReLU hidden layers, and \TreePINN{} uses the sigmoid/ReLU/linear architecture. Wins and ties count paired outcomes at epoch 50.}
\label{tab:init-paired}
\begin{tabular*}{\textwidth}{@{\extracolsep{\fill}}lrrrr@{}}\toprule
Task & Initial difference & Early difference & Epoch-50 difference & Wins / ties\\\midrule
\multicolumn{5}{l}{\textbf{Classification (pp)}}\\
\addlinespace[2pt]
Bank & $+20.6810\pm16.8523$ & $+1.8377\pm1.2416$ & $+1.0130\pm1.0316$ & 17/20, 0\\
Banknote & $+11.0123\pm12.9288$ & $+4.7013\pm4.0139$ & $+0.0000\pm0.0000$ & 0/20, 20\\
Breast Cancer & $+9.8659\pm21.6667$ & $+13.5015\pm7.8578$ & $+1.1462\pm1.3192$ & 14/20, 3\\
Digits & $+42.0414\pm8.9373$ & $+13.5647\pm3.5097$ & $+1.5305\pm1.0771$ & 18/20, 0\\
Spambase & $+22.4238\pm18.5579$ & $+2.1418\pm0.7300$ & $+0.7117\pm1.0580$ & 16/20, 0\\
Wine & $+38.9319\pm18.0852$ & $+58.8483\pm14.0165$ & $+23.5035\pm17.3461$ & 20/20, 0\\
\midrule
\multicolumn{5}{l}{\textbf{Regression (target units)}}\\
\addlinespace[2pt]
California ($10^5$ USD) & $-0.2503\pm0.1939$ & $-0.0703\pm0.0160$ & $-0.0265\pm0.0194$ & 20/20, 0\\
Diabetes (score) & $-26.2528\pm19.8648$ & $-1.8988\pm6.2128$ & $+9.3855\pm5.9931$ & 2/20, 0\\
Airfoil (dB) & $-2.0461\pm0.7169$ & $-1.3008\pm0.3890$ & $-0.5980\pm0.3380$ & 20/20, 0\\
Concrete (MPa) & $-5.6195\pm1.8999$ & $-3.3040\pm0.8744$ & $-0.6060\pm0.5547$ & 16/20, 0\\
Protein (RMSD) & $-1.1222\pm0.3683$ & $-0.1034\pm0.0373$ & $-0.0619\pm0.0765$ & 18/20, 0\\
Gas CO (mg/m$^3$) & $-0.8576\pm0.6233$ & $-0.0751\pm0.0312$ & $-0.0370\pm0.0629$ & 15/20, 0\\
\bottomrule\end{tabular*}
\end{table*}

\clearpage

\clearpage

\begin{table*}[p]\centering\small
\setlength{\tabcolsep}{2.5pt}
\renewcommand{\arraystretch}{1.10}

\caption{Scale and readout ablation at initialization, averaged over epochs 1--10, and at epoch 50. Values are mean $\pm$ sample SD over twenty paired seeds. Classification uses validation Macro-F1 (\%; higher is better), and regression uses validation RMSE in the displayed units (lower is better). Removing adaptive scales fixes every split slope at 20; removing the leaf-informed readout replaces it with a random matrix of equal Frobenius norm. \emph{Without both} applies both changes. Boldface and italics indicate the best and second-best distinct unrounded means within each task and stage. The learning curves additionally report classification cross-entropy.}
\label{tab:init-components}
\begin{tabular*}{\textwidth}{@{\extracolsep{\fill}}lrrrrr@{}}\toprule
Initialization & \shortstack{Digits\\F1 (\%) $\uparrow$} & \shortstack{Spambase\\F1 (\%) $\uparrow$} & \shortstack{Concrete\\RMSE (MPa) $\downarrow$} & \shortstack{Protein\\RMSE (RMSD) $\downarrow$} & \shortstack{Gas CO\\RMSE (mg/m$^3$) $\downarrow$}\\
\midrule
\multicolumn{6}{l}{\textbf{Before optimization}}\\
\addlinespace[2pt]
Without both & $6.879\pm4.557$ & $45.745\pm24.527$ & $17.657\pm1.190$ & $6.154\pm0.134$ & $3.011\pm0.654$\\
Without adaptive scales & $\mathbf{64.624\pm3.202}$ & $\mathbf{88.558\pm1.559}$ & $\mathit{13.895\pm0.647}$ & $\mathit{5.845\pm0.039}$ & $\mathit{1.864\pm0.067}$\\
Without leaf readout & $7.251\pm4.855$ & $52.684\pm14.016$ & $18.162\pm1.463$ & $6.320\pm0.269$ & $3.184\pm0.621$\\
\TreePINN{} & $\mathit{47.154\pm8.980}$ & $\mathit{63.795\pm16.727}$ & $\mathbf{12.261\pm0.508}$ & $\mathbf{5.380\pm0.045}$ & $\mathbf{1.830\pm0.098}$\\
\midrule
\multicolumn{6}{l}{\textbf{Mean over epochs 1--10}}\\
\addlinespace[2pt]
Without both & $77.606\pm2.838$ & $\mathit{93.365\pm0.816}$ & $7.382\pm0.481$ & $4.676\pm0.042$ & $1.250\pm0.090$\\
Without adaptive scales & $\mathit{88.289\pm1.937}$ & $\mathbf{93.508\pm0.848}$ & $\mathit{7.051\pm0.463}$ & $4.678\pm0.047$ & $1.255\pm0.086$\\
Without leaf readout & $82.946\pm2.936$ & $92.981\pm1.015$ & $7.295\pm0.582$ & $\mathbf{4.584\pm0.045}$ & $\mathbf{1.233\pm0.083}$\\
\TreePINN{} & $\mathbf{93.828\pm2.623}$ & $93.216\pm0.945$ & $\mathbf{7.029\pm0.448}$ & $\mathit{4.593\pm0.042}$ & $\mathit{1.241\pm0.081}$\\
\midrule
\multicolumn{6}{l}{\textbf{Epoch 50}}\\
\addlinespace[2pt]
Without both & $94.266\pm1.339$ & $\mathit{93.439\pm0.820}$ & $\mathit{5.518\pm0.418}$ & $4.104\pm0.056$ & $\mathbf{1.143\pm0.073}$\\
Without adaptive scales & $94.466\pm1.201$ & $93.414\pm0.743$ & $5.596\pm0.503$ & $4.109\pm0.071$ & $1.173\pm0.101$\\
Without leaf readout & $\mathit{96.909\pm0.676}$ & $93.309\pm1.213$ & $\mathbf{5.414\pm0.562}$ & $\mathbf{4.021\pm0.085}$ & $1.148\pm0.093$\\
\TreePINN{} & $\mathbf{97.196\pm0.742}$ & $\mathbf{93.441\pm0.965}$ & $5.541\pm0.546$ & $\mathit{4.029\pm0.066}$ & $\mathit{1.146\pm0.107}$\\
\bottomrule
\end{tabular*}
\end{table*}

\begin{table*}[p]\centering\small\setlength{\tabcolsep}{3pt}\renewcommand{\arraystretch}{1.10}
\caption{Corrected statistical comparisons for the scale and readout ablations. Panel (a) reports epoch-50 adjusted $p$-values from the 75-comparison family with twenty paired seeds. The five columns test adaptive scales with leaf-informed and random readouts, leaf-informed readouts with adaptive and fixed scales, and their interaction. Only the adaptive-scale effect on Digits with the leaf-informed readout is significant. Panel (b) reports adjusted $p$-values from the separate twelve-comparison scale study with ten paired seeds. Signs indicate effect direction: positive favors adaptive scales through lower early cross-entropy or higher epoch-50 F1.}\label{tab:ablation-scale-corrected}
\begin{tabular*}{\textwidth}{@{\extracolsep{\fill}}lrrrrr@{}}\toprule
\multicolumn{6}{@{}l}{\textbf{(a) Twenty-seed factorial: epoch 50}}\\
Task & Scale/leaf & Scale/random & Leaf/adaptive & Leaf/fixed & Interaction\\\midrule
Digits & $0.023$ & $0.083$ & $1.000$ & $1.000$ & $1.000$\\
Spambase & $1.000$ & $1.000$ & $1.000$ & $1.000$ & $1.000$\\
Concrete (MPa) & $1.000$ & $1.000$ & $1.000$ & $1.000$ & $1.000$\\
Protein (RMSD) & $1.000$ & $1.000$ & $1.000$ & $1.000$ & $1.000$\\
Gas CO (mg/m$^3$) & $1.000$ & $1.000$ & $1.000$ & $1.000$ & $1.000$\\
\bottomrule\end{tabular*}\par\medskip
\begin{tabular*}{\textwidth}{@{\extracolsep{\fill}}lrr@{}}\toprule
\multicolumn{3}{@{}l}{\textbf{(b) Ten-seed scale study}}\\Task & Early CE benefit: adjusted $p$ & Epoch-50 F1 benefit: adjusted $p$\\\midrule
Bank & $-$; $0.019$ & $-$; $1.000$\\
Banknote & $-$; $0.859$ & $+$; $1.000$\\
Breast Cancer & $+$; $1.000$ & $+$; $1.000$\\
Digits & $+$; $1.64{\times}10^{-5}$ & $+$; $0.014$\\
Spambase & $-$; $1.000$ & $+$; $1.000$\\
Wine & $-$; $0.164$ & $+$; $1.000$\\
\bottomrule\end{tabular*}\label{tab:ablation-readout-corrected}\label{tab:ablation-scale10-corrected}\end{table*}

\clearpage

\clearpage

\begin{table*}[p]\centering\small
\setlength{\tabcolsep}{2.5pt}
\renewcommand{\arraystretch}{1.10}

\caption{Conditional effects of scale calibration and readout initialization at epoch 50: mean $\pm$ sample SD of differences over twenty paired seeds. Scale effects compare adaptive with fixed scales, and readout effects compare leaf-informed with random readouts, with the second setting subtracted from the first. Positive differences favor the first setting for Macro-F1, measured in percentage points (pp); negative differences favor it for RMSE. Each column holds the other factor at the indicated setting.}
\label{tab:init-component-pairs}
\begin{tabular*}{\textwidth}{@{\extracolsep{\fill}}llrrrr@{}}\toprule
Task & Metric & \shortstack{Scale effect\\leaf readout} & \shortstack{Scale effect\\random readout} & \shortstack{Readout effect\\adaptive scales} & \shortstack{Readout effect\\fixed scales}\\\midrule
Digits & F1 pp & $+2.7293\pm1.0211$ & $+2.6425\pm1.1442$ & $+0.2869\pm0.6020$ & $+0.2001\pm0.7996$\\
Spambase & F1 pp & $+0.0276\pm0.9693$ & $-0.1298\pm1.0384$ & $+0.1322\pm0.8937$ & $-0.0252\pm0.6874$\\
\addlinespace[3pt]
Concrete & RMSE (MPa) & $-0.0552\pm0.3159$ & $-0.1040\pm0.2573$ & $+0.1273\pm0.2863$ & $+0.0785\pm0.2626$\\
Protein & RMSE (RMSD) & $-0.0799\pm0.0621$ & $-0.0832\pm0.0831$ & $+0.0080\pm0.0515$ & $+0.0047\pm0.0719$\\
Gas CO & RMSE (mg/m$^3$) & $-0.0265\pm0.0809$ & $+0.0055\pm0.0935$ & $-0.0019\pm0.0941$ & $+0.0301\pm0.0809$\\
\bottomrule
\end{tabular*}
\end{table*}

\begin{table*}[p]
\centering
\small
\setlength{\tabcolsep}{2.5pt}
\renewcommand{\arraystretch}{1.10}

\caption{Adaptive-scale ablation comparing fixed-scale \TreePINN{} with \TreePINN{} over ten paired seeds per task. Cross-entropy (CE; lower is better) is averaged over epochs 1--10 within each seed; Macro-F1 (\%; higher is better) is measured at epoch 50. Values are mean $\pm$ sample SD. Differences are \TreePINN{} minus the fixed-scale variant, with sample SD of paired differences; F1 differences are percentage points. Boldface and italics rank the two condition means within each metric using distinct unrounded values; numerical ties within $10^{-12}$ share rank.}
\label{tab:init-ablation}
\begin{tabular*}{\textwidth}{@{\extracolsep{\fill}}lrrrrrr@{}}
\toprule
Task & \shortstack{Fixed-scale\\ CE} & \shortstack{\TreePINN{}\\ CE} & CE difference & \shortstack{Fixed-scale\\ F1} & \shortstack{\TreePINN{}\\ F1} & F1 difference\\
\midrule
Bank & $\mathbf{0.232\pm0.003}$ & $\mathit{0.240\pm0.004}$ & $0.007\pm0.003$ & $\mathbf{73.29\pm1.35}$ & $\mathit{72.25\pm1.56}$ & $-1.03\pm1.42$\\
Banknote & $\mathbf{0.048\pm0.011}$ & $\mathit{0.078\pm0.022}$ & $0.030\pm0.025$ & $\mathit{99.85\pm0.26}$ & $\mathbf{99.93\pm0.23}$ & $0.07\pm0.38$\\
Breast Cancer & $\mathit{0.178\pm0.043}$ & $\mathbf{0.172\pm0.040}$ & $-0.006\pm0.037$ & $\mathit{94.55\pm1.90}$ & $\mathbf{96.24\pm1.33}$ & $1.69\pm2.27$\\
Digits & $\mathit{0.401\pm0.033}$ & $\mathbf{0.228\pm0.025}$ & $-0.173\pm0.023$ & $\mathit{94.18\pm1.13}$ & $\mathbf{97.17\pm0.93}$ & $2.99\pm0.98$\\
Spambase & $\mathbf{0.180\pm0.022}$ & $\mathit{0.185\pm0.016}$ & $0.005\pm0.009$ & $\mathit{93.21\pm0.74}$ & $\mathbf{93.39\pm1.04}$ & $0.18\pm1.00$\\
Wine & $\mathbf{0.580\pm0.049}$ & $\mathit{0.704\pm0.048}$ & $0.124\pm0.065$ & $\mathit{92.44\pm5.36}$ & $\mathbf{96.82\pm2.44}$ & $4.37\pm5.40$\\
\bottomrule
\end{tabular*}
\end{table*}

\begin{table*}[p]
\centering
\small
\setlength{\tabcolsep}{2.5pt}
\renewcommand{\arraystretch}{1.10}

\caption{Node-assignment diagnostic comparing \TreePINN{} with a slope-permutation control over ten paired seeds. Cross-entropy (CE; lower is better) is averaged over epochs 1--10 within each seed; Macro-F1 (\%; higher is better) is measured at epoch 50. Values are mean $\pm$ sample SD. Differences are \TreePINN{} minus the control, with sample SD of paired differences; F1 differences are percentage points. Permutation within matching signed-replica groups preserves the grouped slope and signed-weight multisets, while threshold-scaled bias multisets generally change. Boldface and italics rank the two condition means within each metric using distinct unrounded values; numerical ties within $10^{-12}$ share rank.}
\label{tab:init-assignment}
\begin{tabular*}{\textwidth}{@{\extracolsep{\fill}}lrrrrrr@{}}
\toprule
Task & Control CE & \shortstack{\TreePINN{}\\ CE} & CE difference & Control F1 & \shortstack{\TreePINN{}\\ F1} & F1 difference\\
\midrule
Digits & $\mathit{0.256\pm0.080}$ & $\mathbf{0.228\pm0.025}$ & $-0.028\pm0.077$ & $\mathit{97.06\pm0.90}$ & $\mathbf{97.17\pm0.93}$ & $0.10\pm0.51$\\
Spambase & $\mathit{0.190\pm0.016}$ & $\mathbf{0.185\pm0.016}$ & $-0.005\pm0.007$ & $\mathit{93.04\pm0.98}$ & $\mathbf{93.39\pm1.04}$ & $0.35\pm1.03$\\
\bottomrule
\end{tabular*}
\end{table*}

\clearpage

\section{Additional Regression Experiments}

Table~\ref{tab:init-reg-exploratory} reports the three-seed experiments on Diamonds, Airbnb Berlin and Superconductor. The table specifies the control initializations used in these experiments, which differ from the controls in the twelve-task benchmark. All entries report validation RMSE at epoch 50 in the corresponding target units. \TreePINN{} has the lowest mean RMSE on Diamonds; Glorot/Tanh leads on Superconductor, and Mixed ReLU leads on Airbnb Berlin.

\section{Source-Tree Transfer and Same-Architecture Control}

\paragraph{Same-architecture random parameters.}
\label{sec:same-architecture-random}
The same-architecture random control retains the \TreePINN{} computation graph, widths, activations, stored parameter count, optimizer and mini-batch order, while replacing every tree-derived affine parameter with an independent draw from the distribution used by the Random baseline. The first layer retains sigmoid gain $g_0=20$; its stored weights and biases are divided by 20 so that $20(Wx+b)$ follows the unscaled random draw. Later layers use the random draw directly. This control removes the transferred thresholds, path identities, adaptive scales and leaf outputs while preserving the architecture. The main Random baseline uses ReLU throughout its hidden layers, and the readout ablation replaces only the output initialization.

The control is evaluated on all twelve main tasks with twenty paired split seeds per task, yielding 240 additional trajectories. Paired runs share training and validation data, parameter counts and mini-batch order; reserved test partitions are not evaluated. Table~\ref{tab:same-architecture-random} reports performance at initialization, averaged over epochs 1--10 and at the primary endpoint of epoch 50. Inference uses the corrected-resampled paired-$t$ approximation with the exact validation-to-training ratio for each task and seed, 19 degrees of freedom, pointwise 95\% intervals, and Holm adjustment across all 36 task--stage comparisons. Win, tie and loss counts summarize the observed paired outcomes.

\newpage

\TreePINN{} improves mean initial performance on all twelve tasks and mean early performance on eleven, with Bank as the exception. At epoch 50, corrected comparisons identify advantages on Airfoil, Concrete and Protein. California and Gas have favorable point estimates without corrected significance, while Diabetes has an adverse point estimate ($p_{\mathrm{Holm}}=.143$). None of the six classification contrasts is statistically significant. The benefits of the complete tree-derived initialization vary across datasets; nonsignificant comparisons do not establish equivalence.

\subsection{Source Trees and Transferred Predictions}

We compare each fitted source tree with its paired network at initialization and after the fixed 50-epoch training schedule, using twenty paired validation splits per main task. A classification tree predicts the majority class in the reached leaf; a regression tree predicts its leaf mean, restored to the original target units using the training-target mean and population SD. Network performance at both stages is taken from the corresponding experimental runs. Tables~\ref{tab:source-tree-classification} and~\ref{tab:source-tree-regression} report all twelve tasks.

Classification transfer agreement is the fraction of validation observations for which the source tree and initialized network predict the same class. For regression, transfer error is $\operatorname{RMSE}(f_0-t)/s_y$, where $f_0$ and $t$ are the initialized-network and tree predictions in the original target units, and $s_y$ is the training-target population SD. These measures evaluate the initialized network directly; the calibration proxy is used for slope selection. The target-based normalization is applied consistently across regression tasks and seeds.

For 220 of the 240 dataset--seed pairs, source-tree performance and transfer error are obtained from fitted trees and initial model states. For California and Diabetes seeds 100--109, initial model states were not retained separately, so the comparison uses source-tree validation metrics and initial transfer errors measured during the original runs and matched to the same trees and inputs. All comparisons use the fitted models and measurements from the main evaluation, without additional tree fitting or neural training. Means and sample SDs summarize the twenty paired validation splits, and test observations are not evaluated.

\clearpage

\begin{table*}[p]\centering\small
\setlength{\tabcolsep}{2.5pt}
\renewcommand{\arraystretch}{1.10}

\caption{Validation RMSE at epoch 50 on three additional regression tasks: mean $\pm$ sample SD over three seeds. These experiments use the following random initializations: Glorot/Tanh uses Xavier-normal hidden weights with gain $5/3$; Mixed ReLU uses He hidden weights and a Xavier-normal output; Same-graph He retains the \TreePINN{} sigmoid first layer and subsequent ReLU layers with He initialization. SF denotes an unavailable result; the DJINN entry is incompatible with the target network dimensions. Boldface and italics indicate the best and second-best distinct unrounded means.}
\label{tab:init-reg-exploratory}
\begin{tabular*}{\textwidth}{@{\extracolsep{\fill}}lrrr@{}}\toprule
Configuration & Superconductor (K) & Diamonds (log target) & Airbnb (USD)\\\midrule
Glorot/Tanh & $\mathbf{12.493\pm0.764}$ & $0.134\pm0.032$ & $201.115\pm31.541$\\
Mixed ReLU & SF & $0.114\pm0.007$ & $\mathbf{199.042\pm32.191}$\\
Same-graph He & $13.445\pm0.314$ & $\mathit{0.110\pm0.006}$ & $\mathit{199.664\pm32.240}$\\
STBI & $13.436\pm0.665$ & $0.116\pm0.004$ & $201.515\pm31.725$\\
DJINN & SF & $0.128\pm0.020$ & $202.164\pm31.806$\\
\TreePINN{} & $\mathit{13.062\pm0.553}$ & $\mathbf{0.099\pm0.004}$ & $201.565\pm31.073$\\
\bottomrule\end{tabular*}
\end{table*}

\begin{table*}[p]
\centering
\small
\setlength{\tabcolsep}{2.5pt}
\renewcommand{\arraystretch}{1.10}

\caption{Comparison with random initialization in the same architecture over twenty paired seeds per task. Positive effects favor \TreePINN{}: classification uses Macro-F1 percentage points (pp), and regression uses the reduction in RMSE in target units. Effects are mean $\pm$ sample SD of paired differences at initialization ($E_0$), averaged over epochs 1--10 within each seed (early), and at epoch 50 ($E_{50}$). Epoch-50 confidence intervals (CI) use the corrected-resampled standard error. W/T/L gives the numbers of wins, ties and losses at epoch 50. $p_{\mathrm{Holm}}$ is the two-sided corrected-resampled paired-$t$ $p$-value after Holm adjustment across all 36 task--stage comparisons.}
\label{tab:same-architecture-random}
\begin{tabular*}{\textwidth}{@{\extracolsep{\fill}}lrrrrrr@{}}
\toprule
Task & \shortstack{$E_0$\\effect} & \shortstack{Early\\effect} & \shortstack{$E_{50}$\\effect} & \shortstack{$E_{50}$ 95\%\\CI} & W/T/L & \shortstack{$p_{\mathrm{Holm}}$}\\
\midrule
\multicolumn{7}{l}{\textbf{Classification (Macro-F1 pp)}}\\
\addlinespace[2pt]
Bank & $19.134\pm17.884$ & $-0.608\pm0.962$ & $0.399\pm1.230$ & $[-1.194,1.993]$ & 14/0/6 & $1.000$ \\
Banknote & $14.623\pm13.693$ & $4.308\pm3.719$ & $0.019\pm0.083$ & $[-0.089,0.126]$ & 1/19/0 & $1.000$ \\
Breast Cancer & $18.742\pm17.820$ & $3.173\pm3.620$ & $0.471\pm0.936$ & $[-0.744,1.686]$ & 10/8/2 & $1.000$ \\
Digits & $44.317\pm9.506$ & $3.097\pm2.659$ & $0.011\pm0.660$ & $[-0.844,0.866]$ & 10/0/10 & $1.000$ \\
Spambase & $26.057\pm19.126$ & $0.314\pm0.398$ & $0.114\pm0.875$ & $[-1.019,1.248]$ & 13/0/7 & $1.000$ \\
Wine & $46.802\pm21.887$ & $53.822\pm21.659$ & $7.541\pm15.563$ & $[-12.792,27.873]$ & 12/4/4 & $1.000$ \\
\midrule
\multicolumn{7}{l}{\textbf{Regression (RMSE reduction in target units)}}\\
\addlinespace[2pt]
California ($10^5$ USD) & $0.447\pm0.279$ & $0.081\pm0.016$ & $0.041\pm0.022$ & $[0.016,0.067]$ & 20/0/0 & $0.068$ \\
Diabetes (score) & $35.088\pm23.572$ & $6.525\pm8.675$ & $-11.136\pm6.640$ & $[-18.771,-3.501]$ & 0/0/20 & $0.143$ \\
Airfoil (dB) & $3.090\pm1.474$ & $1.943\pm0.361$ & $1.113\pm0.491$ & $[0.550,1.677]$ & 20/0/0 & $0.015$ \\
Concrete (MPa) & $8.050\pm3.305$ & $4.298\pm0.869$ & $1.195\pm0.437$ & $[0.693,1.697]$ & 20/0/0 & $0.002$ \\
Protein (RMSD) & $1.725\pm1.137$ & $0.404\pm0.051$ & $0.230\pm0.071$ & $[0.149,0.310]$ & 20/0/0 & $3.14{\times}10^{-4}$ \\
Gas CO (mg/m$^3$) & $0.910\pm0.544$ & $0.118\pm0.040$ & $0.091\pm0.080$ & $[0.000,0.182]$ & 17/0/3 & $0.795$ \\
\bottomrule
\end{tabular*}
\end{table*}

\clearpage

\begin{table*}[p]
\centering
\small
\setlength{\tabcolsep}{2.5pt}
\renewcommand{\arraystretch}{1.10}

\caption{Source-tree and network classification performance on validation data: mean $\pm$ sample SD over twenty paired seeds. $E_0$ denotes the network at initialization, and $E_{50}$ denotes epoch 50. Macro-F1, accuracy and agreement are percentages. Agreement is the fraction of validation observations for which the source tree and initialized network predict the same class and is reported once per task. Higher Macro-F1 and accuracy are better.}
\begin{tabular*}{\textwidth}{@{\extracolsep{\fill}}llrrrr@{}}
\toprule
Task & Metric & Source tree & Network $E_0$ & Network $E_{50}$ & Agreement (\%) \\
\midrule
Bank & Macro-F1 & $63.937\pm7.411$ & $50.616\pm1.882$ & $72.624\pm1.617$ & $88.196\pm9.226$ \\
 & Accuracy & $85.107\pm5.746$ & $88.596\pm0.274$ & $90.123\pm0.245$ & -- \\
\addlinespace[3pt]
Banknote & Macro-F1 & $96.416\pm1.028$ & $49.275\pm11.479$ & $99.963\pm0.165$ & $62.153\pm6.649$ \\
 & Accuracy & $96.460\pm1.019$ & $61.898\pm7.160$ & $99.964\pm0.163$ & -- \\
\addlinespace[3pt]
Breast Cancer & Macro-F1 & $92.470\pm2.162$ & $50.389\pm16.807$ & $96.664\pm1.352$ & $53.991\pm14.814$ \\
 & Accuracy & $92.939\pm2.062$ & $53.246\pm14.150$ & $96.886\pm1.256$ & -- \\
\addlinespace[3pt]
Digits & Macro-F1 & $64.654\pm3.220$ & $47.154\pm8.980$ & $97.196\pm0.742$ & $72.604\pm10.224$ \\
 & Accuracy & $65.696\pm2.934$ & $50.655\pm7.967$ & $97.214\pm0.729$ & -- \\
\addlinespace[3pt]
Spambase & Macro-F1 & $89.580\pm1.314$ & $63.795\pm16.727$ & $93.441\pm0.965$ & $77.989\pm10.157$ \\
 & Accuracy & $90.190\pm1.219$ & $73.321\pm9.527$ & $93.734\pm0.927$ & -- \\
\addlinespace[3pt]
Wine & Macro-F1 & $89.894\pm5.897$ & $64.700\pm19.388$ & $96.784\pm3.098$ & $68.611\pm19.056$ \\
 & Accuracy & $89.861\pm5.860$ & $66.528\pm17.602$ & $96.667\pm3.199$ & -- \\
\bottomrule
\end{tabular*}

\label{tab:source-tree-classification}
\end{table*}

\begin{table*}[p]
\centering
\small
\setlength{\tabcolsep}{2.5pt}
\renewcommand{\arraystretch}{1.10}

\caption{Source-tree and network regression performance on validation data: mean $\pm$ sample SD over twenty paired seeds. $E_0$ denotes the network at initialization, and $E_{50}$ denotes epoch 50. RMSE uses the target units in Table~\ref{tab:init-datasets}, and $R^2$ is dimensionless. Normalized transfer error is $\operatorname{RMSE}(f_0-t)/s_y$, where $f_0$ and $t$ are initialized-network and source-tree predictions on the validation observations, and $s_y$ is the training-target population SD. It is reported once per task. Lower RMSE and higher $R^2$ are better.}
\begin{tabular*}{\textwidth}{@{\extracolsep{\fill}}llrrrr@{}}
\toprule
Task & Metric & Source tree & Network $E_0$ & Network $E_{50}$ & Normalized error \\
\midrule
California & RMSE & $0.718\pm0.015$ & $0.963\pm0.034$ & $0.534\pm0.016$ & $0.592\pm0.048$ \\
 & $R^2$ & $0.610\pm0.014$ & $0.298\pm0.050$ & $0.784\pm0.013$ & -- \\
\addlinespace[3pt]
Diabetes & RMSE & $64.181\pm5.496$ & $62.152\pm4.429$ & $65.435\pm4.858$ & $0.480\pm0.048$ \\
 & $R^2$ & $0.261\pm0.163$ & $0.317\pm0.076$ & $0.231\pm0.173$ & -- \\
\addlinespace[3pt]
Airfoil & RMSE & $4.353\pm0.227$ & $5.157\pm0.380$ & $2.117\pm0.245$ & $0.534\pm0.050$ \\
 & $R^2$ & $0.595\pm0.047$ & $0.432\pm0.072$ & $0.903\pm0.024$ & -- \\
\addlinespace[3pt]
Concrete & RMSE & $8.800\pm0.747$ & $12.261\pm0.508$ & $5.541\pm0.546$ & $0.590\pm0.038$ \\
 & $R^2$ & $0.720\pm0.044$ & $0.459\pm0.039$ & $0.888\pm0.022$ & -- \\
\addlinespace[3pt]
Protein & RMSE & $5.199\pm0.042$ & $5.380\pm0.045$ & $4.029\pm0.066$ & $0.299\pm0.009$ \\
 & $R^2$ & $0.279\pm0.011$ & $0.228\pm0.010$ & $0.567\pm0.015$ & -- \\
\addlinespace[3pt]
Gas CO & RMSE & $1.287\pm0.091$ & $1.830\pm0.098$ & $1.146\pm0.107$ & $0.624\pm0.054$ \\
 & $R^2$ & $0.682\pm0.041$ & $0.358\pm0.064$ & $0.746\pm0.048$ & -- \\
\bottomrule
\end{tabular*}

\label{tab:source-tree-regression}
\end{table*}

\clearpage

\section{Statistical Analysis and Optimization Cost}

\label{sec:statistical-analysis}\label{sec:updates-target-details}
Each comparison pairs a baseline with \TreePINN{} using the same input data, network dimensions and mini-batch schedule. Each seed defines a data split, source tree and initialization, so variation across seeds reflects their combined effects. Early performance is the mean over epochs 1--10 within each seed. Classification uses Macro-F1, and regression uses RMSE in the original target units; the two task types are analyzed separately.

For twenty paired differences $d_s$, we use the corrected resampled approximation~\cite{nadeau1999inference},
\[
 t=\frac{\bar d}{s_d\sqrt{1/20+r}},\qquad
 r=\frac1{20}\sum_s\frac{n_{\mathrm{val},s}}{n_{\mathrm{train},s}},
\]
with sample SD $s_d$ and 19 degrees of freedom. Validation-to-training ratios are approximately $1/3$ for classification and $1/4$ for regression; the calculation uses the actual sample counts for each split. The correction approximates the dependence induced by overlapping splits. Stratification, grouping, coupled sources of randomness and reuse of validation data also affect this dependence, so the resulting inference remains approximate. Pointwise 95\% intervals use the same corrected standard error. When all paired differences are zero, we assign $p=1$ and a zero-width observed interval; this convention describes the observed differences and does not establish equivalence.

Holm adjustment sorts raw values $p_{(i)}$ within each comparison family and computes $\min\{1,\max_{j\le i}(m-j+1)p_{(j)}\}$. The main family contains all five baselines, twelve tasks and three stages ($m=180$; Table~\ref{tab:corrected-main}); the optimization-cost family contains all five baselines and twelve tasks ($m=60$; Table~\ref{tab:corrected-efficiency}). Statistical significance is assessed at adjusted $p<.05$, with adjusted values capped at 1. These retrospective analyses use comparison families defined after descriptive results were available and before adjusted tests were calculated. All 74 significant main comparisons favor \TreePINN{}: 23 at initialization, 39 over epochs 1--10 and 12 at epoch 50. No epoch-50 comparison against He or Xavier is significant. Separate sign tests of effect directions across the six datasets within each task type~\cite{demsar2006statistical} yield no significant results after correction across 30 comparisons.

\newpage

The twenty-seed scale/readout factorial uses a 75-comparison family comprising five tasks, five contrasts and three stages. The separate ten-seed classification scale study uses twelve comparisons: early cross-entropy and epoch-50 F1 on six tasks. Table~\ref{tab:ablation-scale-corrected} reports their adjusted values and effect directions. At epoch 50, only the benefit of adaptive scales on Digits with the leaf-informed readout is significant under the 75-comparison correction; no leaf-readout effect or interaction is significant. In the ten-seed study, adaptive scales significantly increase early cross-entropy on Bank and improve both early cross-entropy and epoch-50 F1 on Digits. Each ablation family is corrected separately from the main comparisons.

\paragraph{Shared performance target and censoring.}
For each of the twelve tasks and twenty paired seeds, all six methods use the He baseline's epoch-50 validation score as their common target. Attainment requires Macro-F1 at or above this target or RMSE at or below it, with numerical tolerance $10^{-10}$. All 1,440 trajectories are evaluated at initialization and at each epoch end. The first observed attainment determines the update count, even if later performance no longer meets the target. Classification uses batches of 128 and regression batches of 64, including a final batch containing any remaining observations. Paired runs use identical batch schedules. Runs that do not reach the target are right-censored at epoch 50. Table~\ref{tab:updates-target-attainment} reports attainment counts and Kaplan--Meier median update counts using all twenty runs per method and task. Fewer than half of the \TreePINN{} runs on Diabetes attain the target, so its median is not reached within the training budget. By construction, every He run attains its own endpoint target.

Budgeted optimization cost is the number of updates performed until first observed attainment or exhaustion of the 50-epoch budget. Runs that do not reach the target incur the full budget and remain nonattainments. Statistical tests compare this bounded update count; Kaplan--Meier medians summarize censored attainment times separately. Fifteen of the sixty cost comparisons significantly favor \TreePINN{}. On Diabetes, \TreePINN{} uses a larger mean fraction of the update budget than each baseline, with no significant corrected cost differences. The update count measures neural optimization; tree fitting and initialization incur additional computation, and elapsed runtime is not measured by this count. All evaluations use validation data; test partitions are not evaluated.

\clearpage

\begin{table*}[p]\centering\small\setlength{\tabcolsep}{3pt}\renewcommand{\arraystretch}{1.10}
\caption{Holm-adjusted two-sided corrected-resampled $p$-values comparing \TreePINN{} with each baseline. All 180 comparisons form one family. $E_0$ denotes initialization, $E_{1:10}$ the within-seed mean over epochs 1--10, and $E_{50}$ epoch 50. All significant differences favor \TreePINN{}; nonsignificant differences do not establish equivalence.}\label{tab:corrected-main}
\begin{tabular*}{\textwidth}{@{\extracolsep{\fill}}llrrrrr@{}}\toprule
Task & Stage & Random & He & Xavier & STBI & DJINN\\\midrule
Bank & $E_0$ & 1 & 1 & 1 & 1 & 1\\
Bank & $E_{1:10}$ & 1 & 1 & 1 & 1 & 1\\
Bank & $E_{50}$ & 1 & 1 & 1 & 1 & 1\\
\addlinespace[2pt]
Banknote & $E_0$ & 1 & 1 & 1 & 1 & 1\\
Banknote & $E_{1:10}$ & 1 & 1 & 1 & 0.0021 & 0.0003\\
Banknote & $E_{50}$ & 1 & 1 & 1 & 1 & 1\\
\addlinespace[2pt]
Breast Cancer & $E_0$ & 1 & 1 & 1 & 1 & 1\\
Breast Cancer & $E_{1:10}$ & 0.2491 & 0.9598 & 1 & $<10^{-4}$ & $<10^{-4}$\\
Breast Cancer & $E_{50}$ & 1 & 1 & 1 & 1 & 1\\
\addlinespace[2pt]
Digits & $E_0$ & $<10^{-4}$ & $<10^{-4}$ & $<10^{-4}$ & $<10^{-4}$ & $<10^{-4}$\\
Digits & $E_{1:10}$ & $<10^{-4}$ & 0.0007 & 0.0001 & $<10^{-4}$ & $<10^{-4}$\\
Digits & $E_{50}$ & 1 & 1 & 1 & 0.0003 & 0.0454\\
\addlinespace[2pt]
Spambase & $E_0$ & 1 & 1 & 1 & 0.8179 & 0.3576\\
Spambase & $E_{1:10}$ & 0.0358 & 0.0164 & 0.2108 & 1 & 0.0082\\
Spambase & $E_{50}$ & 1 & 1 & 1 & 1 & 1\\
\addlinespace[2pt]
Wine & $E_0$ & 0.8947 & 0.2691 & 0.3800 & 0.4110 & 0.0980\\
Wine & $E_{1:10}$ & 0.0041 & 0.0003 & 0.0004 & $<10^{-4}$ & $<10^{-4}$\\
Wine & $E_{50}$ & 1 & 1 & 1 & 1 & 0.0082\\
\addlinespace[2pt]
California & $E_0$ & 0.0592 & 1 & $<10^{-4}$ & $<10^{-4}$ & 0.0635\\
California & $E_{1:10}$ & 0.0003 & $<10^{-4}$ & $<10^{-4}$ & 1 & $<10^{-4}$\\
California & $E_{50}$ & 0.8022 & 1 & 1 & 0.0366 & $<10^{-4}$\\
\addlinespace[2pt]
Diabetes & $E_0$ & 0.8910 & 1 & 0.0017 & 0.0049 & 1\\
Diabetes & $E_{1:10}$ & 1 & 1 & 1 & 1 & 0.4538\\
Diabetes & $E_{50}$ & 0.3837 & 0.8384 & 0.6639 & 0.4794 & 0.4110\\
\addlinespace[2pt]
Airfoil & $E_0$ & 0.0056 & 0.0061 & $<10^{-4}$ & $<10^{-4}$ & 1\\
Airfoil & $E_{1:10}$ & $<10^{-4}$ & 0.0010 & $<10^{-4}$ & $<10^{-4}$ & $<10^{-4}$\\
Airfoil & $E_{50}$ & 0.0363 & 0.4110 & 0.1884 & 0.0039 & $<10^{-4}$\\
\addlinespace[2pt]
Concrete & $E_0$ & 0.0025 & 0.0041 & $<10^{-4}$ & $<10^{-4}$ & 0.0150\\
Concrete & $E_{1:10}$ & $<10^{-4}$ & 0.0002 & $<10^{-4}$ & 0.0349 & $<10^{-4}$\\
Concrete & $E_{50}$ & 1 & 1 & 1 & 0.5113 & 0.0059\\
\addlinespace[2pt]
Protein & $E_0$ & 0.3576 & 0.0029 & $<10^{-4}$ & $<10^{-4}$ & 0.3112\\
Protein & $E_{1:10}$ & $<10^{-4}$ & 0.0083 & $<10^{-4}$ & $<10^{-4}$ & $<10^{-4}$\\
Protein & $E_{50}$ & 0.0164 & 1 & 1 & 0.0025 & $<10^{-4}$\\
\addlinespace[2pt]
Gas CO & $E_0$ & 0.6340 & 1 & $<10^{-4}$ & $<10^{-4}$ & 0.6965\\
Gas CO & $E_{1:10}$ & 0.2733 & 0.0345 & 0.0144 & 1 & 0.0235\\
Gas CO & $E_{50}$ & 1 & 1 & 1 & 1 & 1\\
\addlinespace[2pt]
\bottomrule\end{tabular*}\end{table*}

\clearpage

\begin{table*}[p]\centering\small\setlength{\tabcolsep}{2pt}
\caption{Target attainment and Kaplan--Meier median optimizer updates.}\label{tab:updates-target-attainment}
\noindent\textbf{Classification}\par\smallskip
\begin{tabular*}{\textwidth}{@{\extracolsep{\fill}}lrrrrrrrrrrrr@{}}\toprule
Method & \multicolumn{2}{c}{Bank} & \multicolumn{2}{c}{Banknote} & \multicolumn{2}{c}{Breast Cancer} & \multicolumn{2}{c}{Digits} & \multicolumn{2}{c}{Spambase} & \multicolumn{2}{c}{Wine}\\
\cmidrule(lr){2-3}\cmidrule(lr){4-5}\cmidrule(lr){6-7}\cmidrule(lr){8-9}\cmidrule(lr){10-11}\cmidrule(lr){12-13}
 & $n$/20 & KM & $n$/20 & KM & $n$/20 & KM & $n$/20 & KM & $n$/20 & KM & $n$/20 & KM\\\midrule
Random & 19 & 1060 & 19 & 91 & 17 & 45 & 15 & 144 & 18 & 154 & 10 & 49\\
He & 20 & 1696 & 20 & 133 & 20 & 69 & 20 & 252 & 20 & 154 & 20 & 48\\
Xavier & 19 & 1272 & 18 & 98 & 18 & 42 & 16 & 126 & 19 & 132 & 12 & 44\\
STBI & 18 & 1484 & 8 & NR & 8 & NR & 0 & NR & 19 & 44 & 11 & 44\\
DJINN & 15 & 2120 & 16 & 182 & 18 & 84 & 0 & NR & 18 & 176 & 3 & NR\\
\TreePINN{} & 20 & 848 & 20 & 42 & 19 & 9 & 19 & 45 & 20 & 44 & 20 & 1\\
\bottomrule\end{tabular*}\par\medskip
\noindent\textbf{Regression}\par\smallskip
\begin{tabular*}{\textwidth}{@{\extracolsep{\fill}}lrrrrrrrrrrrr@{}}\toprule
Method & \multicolumn{2}{c}{California} & \multicolumn{2}{c}{Diabetes} & \multicolumn{2}{c}{Airfoil} & \multicolumn{2}{c}{Concrete} & \multicolumn{2}{c}{Protein} & \multicolumn{2}{c}{Gas CO}\\
\cmidrule(lr){2-3}\cmidrule(lr){4-5}\cmidrule(lr){6-7}\cmidrule(lr){8-9}\cmidrule(lr){10-11}\cmidrule(lr){12-13}
 & $n$/20 & KM & $n$/20 & KM & $n$/20 & KM & $n$/20 & KM & $n$/20 & KM & $n$/20 & KM\\\midrule
Random & 11 & 10350 & 19 & 45 & 10 & 800 & 10 & 407 & 3 & NR & 13 & 9452\\
He & 20 & 7866 & 20 & 35 & 20 & 688 & 20 & 418 & 20 & 18320 & 20 & 7506\\
Xavier & 20 & 7038 & 19 & 45 & 14 & 640 & 18 & 330 & 11 & 22442 & 11 & 13066\\
STBI & 10 & 9315 & 17 & 25 & 8 & NR & 7 & NR & 2 & NR & 14 & 6394\\
DJINN & 1 & NR & 18 & 115 & 0 & NR & 0 & NR & 0 & NR & 6 & NR\\
\TreePINN{} & 20 & 2277 & 9 & NR & 20 & 240 & 18 & 132 & 19 & 16030 & 20 & 2502\\
\bottomrule\end{tabular*}\par\medskip
\par\smallskip\begin{minipage}{\textwidth}\small For each method and task, the table reports the number of runs reaching the shared target (out of 20) and the Kaplan--Meier (KM) median optimizer update count, including right-censored runs. NR indicates that the median is not reached within 50 epochs. Each target is the corresponding seed's He validation score at epoch 50; batch schedules are paired across methods.\end{minipage}
\end{table*}

\begin{table*}[p]\centering\small\setlength{\tabcolsep}{3pt}\renewcommand{\arraystretch}{1.10}
\caption{Holm-adjusted $p$-values for budgeted optimization cost, comparing \TreePINN{} with each baseline across one family of 60 comparisons. Cost counts updates until first observed target attainment or exhaustion of the 50-epoch budget. All significant differences favor \TreePINN{}.}\label{tab:corrected-efficiency}
\begin{tabular*}{\textwidth}{@{\extracolsep{\fill}}lrrrrr@{}}\toprule
Task & Random & He & Xavier & STBI & DJINN\\\midrule
Bank & 1 & 1 & 1 & 1 & 1\\
Banknote & 1 & 0.1810 & 1 & 0.0765 & 0.1810\\
Breast Cancer & 1 & 0.4230 & 1 & 0.8556 & 0.1229\\
Digits & 1 & 0.8556 & 1 & 0.0002 & 0.0002\\
Spambase & 1 & 1 & 1 & 1 & 1\\
Wine & 0.0177 & $<10^{-4}$ & 0.2808 & 0.0620 & $<10^{-4}$\\
\midrule
California & 0.0203 & 0.0045 & 0.0346 & 0.1368 & $<10^{-4}$\\
Diabetes & 1 & 1 & 1 & 1 & 1\\
Airfoil & 0.0004 & 0.0012 & 0.0305 & $<10^{-4}$ & $<10^{-4}$\\
Concrete & 0.1298 & 0.3685 & 0.7389 & 0.6215 & 0.0395\\
Protein & 0.0875 & 1 & 0.1368 & 0.0644 & 0.1229\\
Gas CO & 0.3052 & 0.7083 & 0.4282 & 0.4718 & 0.1368\\
\bottomrule\end{tabular*}\end{table*}

\clearpage

\end{document}